\documentclass[11pt, a4paper, twocolumn]{article}
\usepackage[utf8]{inputenc}
\usepackage{multicol}
\usepackage{fancyhdr}
\usepackage{setspace}
\usepackage{indentfirst}
\usepackage{amsmath}
\usepackage{graphicx}
\usepackage{caption}
\usepackage[subrefformat=parens,labelformat=parens]{subcaption}
\usepackage[a4paper]{geometry}
\usepackage{hyperref}
\usepackage[backend=bibtex,style=numeric-comp,sorting=none,firstinits=true,maxbibnames=99]{biblatex}
\DeclareNameAlias{author}{last-first}
\bibliography{references}

\usepackage{float}
\usepackage{tabularx}

\usepackage{sectsty} 
\sectionfont{\Large} 
\subsectionfont{\large} 

\usepackage{lipsum}  

\makeatletter
\renewcommand{\maketitle}{\bgroup\setlength{\parindent}{0pt}
\begin{flushleft}
  \onehalfspacing
  \fontsize{20}{23}\selectfont
  \textbf{\@title} \\
  \hfill \break
  \fontsize{12}{13}\selectfont
  \@author
\end{flushleft}\egroup
}
\makeatother

\title{A data filling methodology for time series based on CNN and (Bi)LSTM neural networks}
\author{%
        \textbf{Kostas Tzoumpas$^{1}$, Aaron Estrada$^{1}$, Pietro Miraglio$^{2,*}$, Pietro Zambelli$^{1}$}\\
        \fontsize{11}{14}\selectfont
        $^{1}$ Eurac Research – Institute for Renewable Energy, Bolzano, Italy \\
        $^{2}$ Centro Euro-Mediterraneo sui Cambiamenti Climatici, Bologna, Italy \\
        Email: $^{1}$mail@ktzoympas.anonaddy.com, \{aaron.estrada, pietro.zambelli\}@eurac.edu, $^{2}$pietro.miraglio@cmcc.it ($^{*}$corresponding author)
        }

\begin{document}

\twocolumn[
  \begin{@twocolumnfalse}
    \maketitle

%
%
\noindent \textbf{Abstract.} In the process of collecting data from sensors, several circumstances can affect their continuity and validity, resulting in alterations of the data or loss of information. Although classical methods of statistics, such as interpolation-like techniques, can be used to approximate the missing data in a time series, the recent developments in Deep Learning (DL) have given impetus to innovative and much more accurate forecasting techniques. In the present paper, we develop two DL models aimed at filling data gaps, for the specific case of internal temperature time series obtained from monitored apartments located in Bolzano, Italy.
The DL models developed in the present work are based on the combination of Convolutional Neural Networks (CNNs), Long Short-Term Memory Neural Networks (LSTMs), and Bidirectional LSTMs (BiLSTMs). Two key features of our models are the use of both pre- and post-gap data, and the exploitation of a correlated time series (the external temperature) in order to predict the target one (the internal temperature).
Our approach manages to capture the fluctuating nature of the data and shows good accuracy in reconstructing the target time series. 
In addition, our models significantly improve the already good results from another DL architecture that is used as a baseline for the present work. \\

\noindent \textbf{Keywords:}  data ﬁlling, 
time series, 
sensor data, 
LSTM, 
Neural Networks.

  \end{@twocolumnfalse}
  \vspace{1.5em}
]
\thispagestyle{empty}


%
%
%
%
%
\section{Introduction} \label{introduction}
A time series is a sequence of discrete-time data in which the information is indexed at successive points in time. Data sequences of this kind are largely used in Finance, Natural Sciences, Humanities, and in general in any domain in which some quantity can be repeatedly measured over time.
During the last decades, the ability to analyze time series has grown significantly; on the one hand due to the increasingly higher availability of data in multiple contexts, on the other hand thanks to the advances in Computer Science, both in terms of algorithms and computational power. Nowadays, the motivations for analyzing time series and for trying to forecast their future values can be diverse, including the search for trends in the data and the modelling of natural or human phenomena.

The collection of time series data can be done by several means and the interval between subsequent measurements can vary a lot depending on the context and goal. When granular and reliable monitoring of some variables is required, like in the case of air temperature or wind speed, the recording is made through sensors. 
Leaving aside the quality and accuracy of such sensors, some external factors, like power outages or failures of the data transmission systems, can hinder the data collection. These incidents can cause data losses, leading to impaired time series that cannot be used in deeper analysis due to the lack of continuity in the data stream. 
For these reasons, the development of new algorithms able to forecast or reconstruct the missing data in a time series has become a relevant research topic in the last few decades. 

In this paper, we deal with a data filling problem for internal temperature time series collected by sensors in monitored buildings, where measurements are taken with a 5-minute granularity.
Our approach is based on the use of cutting-edge Deep Learning techniques, that we combine in an original fashion in order to produce two different and independent data filling models.
Our two models, one based on CNNs and LSTMs and the other one on CNNs and Bi-LSTMs, show better data filling capabilities than a model developed in~\cite{song2020time} that we take as a baseline. Besides, in cases where the prediction of the missing values is inaccurate, our models still succeed in identifying the general trend of the series.

It is important to stress that the performances of our models are linked to the context in which we operate (temperature sensors in monitored apartments) and that the model we take as a baseline was developed for filling data gaps in time series recorded in a different sphere.
Nevertheless, in analogy with the DL algorithms for image recognition, it is legitimate to expect that DL models developed to reconstruct time series in a specific framework can achieve good results when trained with time-indexed sequences coming from another context. As we are going to describe, this is indeed what happens with the model we take as a baseline.
For what concerns our work instead, the models we present can be applied in any domain in which a target time series is reconstructed using information coming from the time series itself and from another time series, correlated with the target one, that has no gaps.

The rest of the paper is structured as follows: in Section~\ref{state of the art} we describe the problem of data filling and time series forecasting, introduce the DL network that we use, and outline the model that we take as a baseline; in Section~\ref{methodology} we focus on the details of our framework, describing the methodology we followed for the data preparation and the architecture of the models we developed; in Section~\ref{Results and discussion} we present the results of our two models compared to the baseline; finally, in Section~\ref{Conclusion and future work} we present our conclusions and some possible future lines of research.

%
%
%

\section{State of the art} \label{state of the art}

\subsection{Data filling and time series forecasting}

Broadly speaking, data filling consists of reconstructing the missing values in a data set based on the available data. The nature of the data to be reconstructed is problem dependent and in many cases the filling process is based on features which are not time-indexed, like for instance in the reconstruction of corrupted satellite images. Thus, data filling is not necessarily related to predicting future values, but those two notions are strictly related in the case of time series that we treat in this paper. Indeed, the problem of filling a gap in a time series is very similar to the forecasting of the same series, with the important difference that for data filling we have available also the data following, and not only preceding, the values we want to predict. 
Indeed, the most advanced approaches to fill time series use both forward and backward in time forecasting to obtain their final results as a combination of the two previsions, like done for instance in~\cite{song2020time}. This idea, compared to using only the forward in time forecast of the missing values, brings the great advantage of using more information to fill the data gap. In this paper, we follow this more advanced approach, proposing novel ways of combining forward and backwards predictions in a single data filling model.

In time series forecasting, an essential distinction is made between one- and multi-step prediction algorithms.
Essentially, a one-step prediction algorithm creates one future predicted value at a time, while a multi-step prediction algorithm predicts multiple steps at once. 
In the case of filling a gap of more than one value, multi-step prediction algorithms predict all the values of the data gap in a single prediction, using only the originally available data. In contrast, one-step prediction algorithms forecast one value at a time and use already predicted data to proceed and fill the whole gap, resulting in the risk of propagating errors throughout the data gap. 
The models we present in this paper belong to the class of multi-step prediction algorithms, which are considered more advanced and reliable than one-step prediction.

In the context of multi-step predictions, the Machine Learning and DL algorithms have been recently shown to perform better than other more traditional techniques, such as Generalized Regression Neural Networks, Gaussian Processes, or AutoRegressive Integrated Moving Average (ARIMA)~\cite{atiya1999comparison, ahmed2010empirical, cai2019day}.
In particular, among all different DL techniques, the most widely used in time series forecasting are currently the Long Short-Term Memory Neural Networks (LSTM), which were introduced in the influential paper~\cite{hochreiter1997long} --- see~\cite{livieris2020cnn, kim2019predicting} for some applications. 
This kind of neural networks has recently been successfully applied also to data filling problems coming from a variety of fields.
Indeed, in~\cite{song2020time} the authors use forward and backwards in time LSTM models to forecast the missing sequences in a biological context, obtaining their final output as a linear combination of the forward and backward predictions.
In another paper~\cite{ren2019using}, the authors use a single (and forward in time) stacked model of LSTMs for filling multi-dimensional time series from hydrological monitoring. 
In the context of oceanography, the authors of ~\cite{contractor2021efficacy} propose three different models based on LSTM networks aimed at filling data gaps in time series.
Those three models are used for gaps of different lengths: the first two run  standalone to reconstruct data gaps shorter than a certain threshold, 
while the third one is used to fill longer gaps of the time series, making use of the outputs of the first two models. In practice, 
this third model is trained on longer time sequences that consists of data processed by the other two models. 
An advantage of this technique is that it allows to deal with longer gaps. However, it has the drawback of accumulating errors from the first two models. 
Although their models "preserve [...] the seasonal or shorter timescale variability", the authors recognize that they "struggle to reproduce even shorter timescale variability present in the observations"~\cite[p.~11]{contractor2021efficacy}.

All the works we mentioned so far propose DL models composed mainly of LSTM layers. Instead, the recent review paper~\cite{lara2021experimental} highlights how combinations of LSTMs and Convolutional Neural Networks (CNNs) can outperform individual models of LSTMs or CNNs.
Our work follows this promising line of research, integrating CNN and LSTM architectures in two new DL models. 

The first of our models, to which we refer as CNN-LSTM (see Subsection~\ref{CNN-LSTM model}), consists of a forward in time DL network combined with a backwards in time one. The two networks have identical structure but are trained on data respectively preceding and following the data gap. The final output of the model is then a combination of the two networks' outputs.  
The second model presented in this paper, to which we refer as CNN-BiLSTM (see Subsection~\ref{CNN-BiLSTM model}), consists instead of CNN and BiLSTM layers combined in a single DL architecture. Thus, once trained, it outputs its predictions in a single step, and not as a combination of two sequences.
We were motivated in developing this second model by several recent studies~\cite{rai2020cnn, le2019improving, hong2020multivariate} suggesting that CNN-BiLSTM substantially improve other DL alternatives such as CNN-LSTMs in the context of time series forecasting.


In the present paper, we choose as a baseline an LSTM model developed in~\cite{song2020time}, which was shown to perform largely better than traditional methods in the case of highly fluctuating and nonlinear data, as is the case of the internal temperature time series we treat. 
Indeed, both works~\cite{song2020time} and~\cite{ren2019using} mentioned above compare their results with the more traditional AutoRegressive Integrated Moving Average (ARIMA), which, as it is stressed in~\cite[p.~14]{song2020time}, is considered to be in this field "the best among several traditional methods".
The results in~\cite{song2020time, ren2019using} make evident that LSTM-based neural networks outperform ARIMA when there are high nonlinearities and several fluctuations in the data to be filled.

\subsection{Deep Learning architectures} \label{Deep Learning architectures}

This section is a concise presentation of the Deep Learning architectures used in this paper. Without the claim of being exhaustive, it aims at providing an outline of their structure and functioning before applying them to our specific case.
Specifically, the deep networks we use are the Convolutional Neural Networks (CNN), the Long Short-Term Memory Neural Networks (LSTM), and their Bidirectional form (BiLSTM). CNN and LSTM layers are used in the first model presented in this work (see Section~\ref{CNN-LSTM model}), while the second model uses CNN and BiLSTM layers (see Section~\ref{CNN-BiLSTM model}).

\smallskip
\subsubsection*{Convolutional Neural Networks (CNN)} \label{CNN}
CNNs are a class of neural networks able to detect simple and also more complex patterns in the data, often used in processing multidimensional inputs.
They were originally developed to recognize visual patterns in images (lines, curves, but also objects or parts of), becoming quickly one of the most important tools in computer vision and particularly in image recognition~\cite{kuo2016understanding}.  
Moreover, CNNs normally reduce the number of parameters necessary to build up the model, which is another advantage of using them in a network, especially in the first part.

\begin{figure}[h!]
    \centering
    \includegraphics[width=0.8\linewidth]{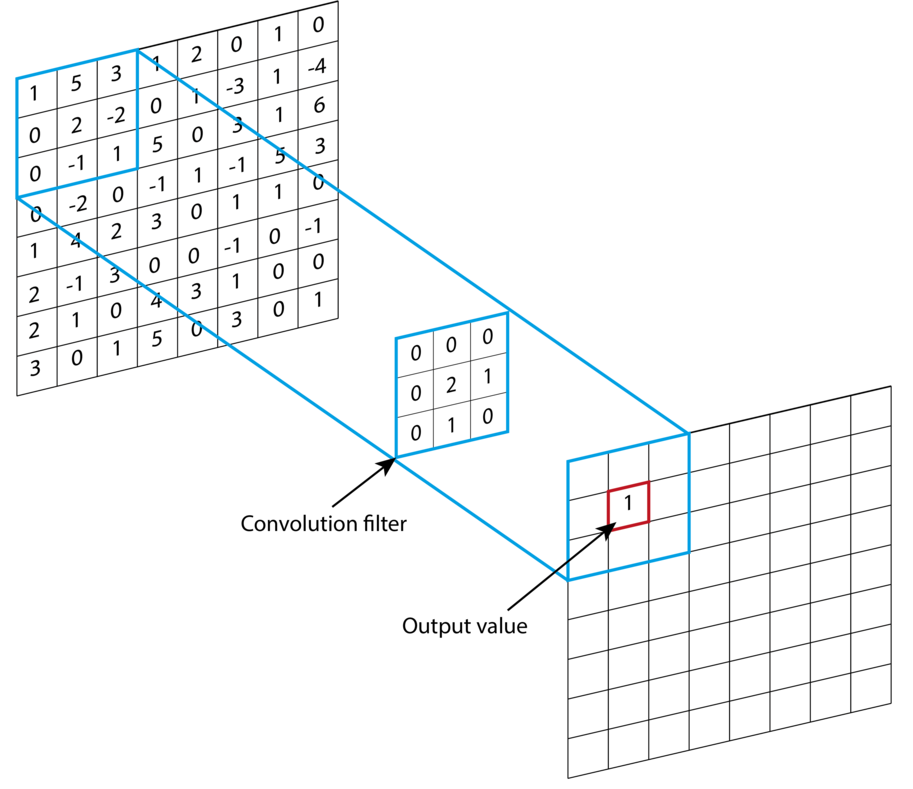}
    \caption{A representation of a convolution operation with a kernel of size 3x3.}
    \label{fig:convolution1}
\end{figure}

The core of a CNN architecture consists of the convolution layers, that perform a convolution operation on the input data and transmit the output to the pooling layers (that are essentially downsampling). 
A convolution is an operation that takes a sliding portion of the input, multiplies it element-wise with a matrix named \textit{Convolution filter} (or simply \textit{kernel}), and sums the elements. The output is a single value for every sample of the input matrix, as shown in Fig.~\ref{fig:convolution1}.

\subsubsection*{Long Short-Term Memory Neural Networks (LSTM)} \label{LSTM}
LSTMs belong to the wide category of Recurrent Neural Networks (RNN), whose key feature is the ability of using loops to keep part of the processed information inside the network. 
Networks of this type are thus able to use information from prior experiences to predict future events, allowing the detection of order dependency in sequences of data. This is why RNNs are employed in several areas regarding data sequences, such as speech recognition, text generation, machine translation, and time series forecasting. 
Though, one of the main disadvantages of the RNNs is that they suffer from the vanishing gradient problem~\cite{bengio1994vanishing, hochreiter1998vanishing}. 
Briefly, it consists of the gradient of the loss function being the product of numerous small terms and thus possibly tending to zero. This is due to the chain rule of calculus and the fact that the loss function depends on all the neurons preceding that point. If the gradient tends to zero, since the model training is based on the update of the weights via the gradient itself, the model is stuck and does not improve.

\begin{figure}[ht]
    \centering
    \includegraphics[width=0.6\linewidth]{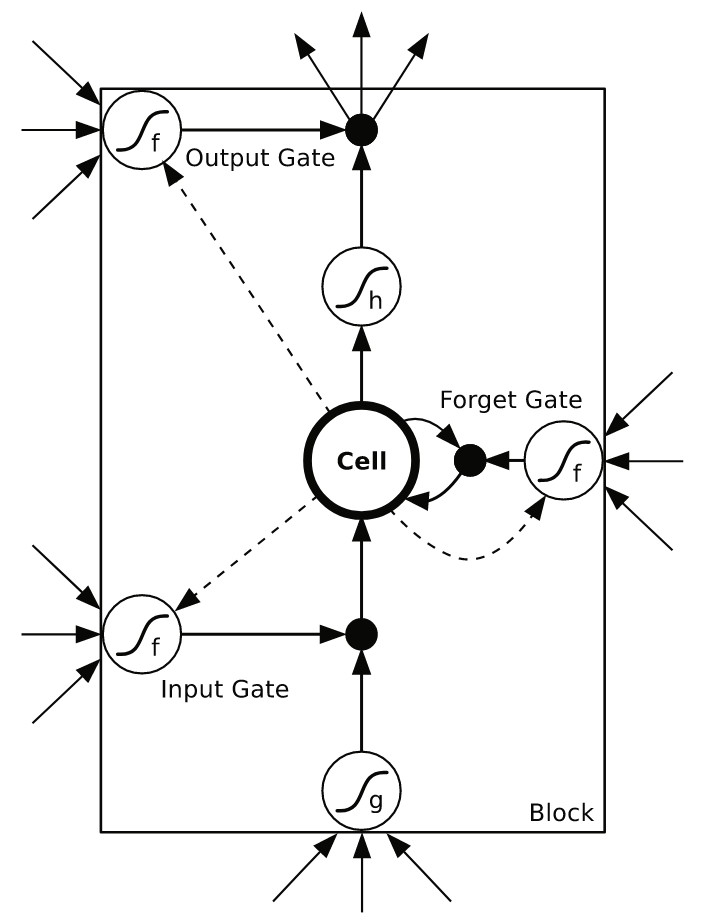}
    \caption{A simple case of LSTM memory block with one cell, as shown in ~\cite{graves2012supervised}.}
    \label{fig:lstm_gates}
\end{figure}

LSTMs were proposed in the seminal paper~\cite{hochreiter1997long} as an improvement of RNNs.  An LSTM cell has three gate units: a memory gate, a forget gate, and an output gate.
Those gates behave like the nodes of a neural network, blocking or passing on information based on a filtering by their own weights, which are modified during the learning process of the network. 
One of the main advances compared to RNNs is in the ability of LSTMs to maintain the information that is propagated over time, preserving a more stable error and overcoming the vanishing gradient problem.
Since it would be difficult to visualise an LSTM network with multiple layers, weights, and gates, Fig.~\ref{fig:lstm_gates} simply shows the positioning of the contents inside an LSTM memory block with one cell~\cite{graves2012supervised}.

\subsubsection*{Bidirectional Long Short-Term Memory Neural Networks (BiLSTM)} \label{BiLSTM}
Bidirectional Recurrent Neural Networks (BRNN) were introduced in~\cite{schuster1997bidirectional} as special RNNs processing the sequence of data in both directions. They inherit all the characteristics of RNNs and extend their capabilities to contexts in which past and future information affects the output, such as natural language processing, sentence completion, translation, relation classification in semantics, and video captioning~\cite{bin2016bidirectional, zhang2015bidirectional}.

A BiLSTM network is thus a combination of a Bidirectional neural layer and an LSTM layer. That is, the LSTM is applied within the Bidirectional layer, resulting in two hidden layers, one for the Forward LSTM and one for the Backward LSTM~\cite{fu2019multi}. 
As a consequence, the output of a BiLSTM network contains twice as many neurons as an LSTM layer with the same input. 
Then, the outputs are combined depending on the user's preference and the task to solve. 
Normally, the results from the forward and backward LSTM are simply concatenated or merged by either multiplication, addition, or their average.
The key characteristic of BiLSTMs is that "for every point in a given sequence, the network has complete, sequential information about all points before and after it"~\cite[p.~74]{zhang2015bidirectional}.

\subsection{Baseline model} \label{Baseline model}

In what follows, we briefly describe the data-filling models developed in~\cite{song2020time}, focusing in particular on the one that we take as a baseline for the present work.

In~\cite{song2020time}, the authors deal with a data-filling problem for time series in a biological context.
More precisely, the time series for which they want to fill data gaps contain the values of stem moisture of a particular species of plants. 
Besides this information, they also have time series of other environmental parameters, which are the soil temperature, air humidity, and photo-synthetically active radiation. 
This additional information is taken into consideration since the authors show there is a high correlation with the plant stem moisture, which is the target of their data-filling work.

All the DL models developed in~\cite{song2020time} have an LSTM architecture at their core.
This neuronal structure is shown in detail in Table~\ref{table:Baseline_parameters}, and consists of 
an LSTM layer of 256 neurons,
a Dropout layer,
two consecutive LSTM layers of 128 and 64 neurons respectively, and 
a Fully Connected (Dense) layer. 

\begin{table}[h!]
    \centering
    \caption{The layers and trainable parameters of the LSTM network in~\cite{song2020time}.}
    \label{table:Baseline_parameters}
    \begin{tabular}{c c c} 
         Layers & Neurons & Parameters \\ [1ex]
         \hline\hline \\
         LSTM 1 & 256 & 264,192 \\ 
         Dropout & - & 0 \\
         LSTM 2 & 128 & 197,120 \\
         LSTM 3 & 64 & 49,408 \\
         Dense & 1 & 65 \\ [1ex]
         \hline\hline \\ [1ex]
         Total parameters & & 510,785 \\ [1ex]
         \hline\hline \\
    \end{tabular}
\end{table}

In~\cite{song2020time}, the authors develop several data-filling models that can be grouped into two different categories, depending on the type of input data they use in order to fill the gaps in the target time series. 
The first group of methods, called "One-dimensional", uses only the target time series for the prediction of its missing values. 
The data-filling capability is restricted to gaps of about one day, but the results are much better than the ones obtained by traditional data filling methods, such as ARIMA, that do not use Deep Learning techniques. 
For filling longer gaps, these "One-dimensional" methods would need to be applied several times on sliding 1-day windows, using already predicted values for subsequent predictions. 
This procedure would accumulate the errors and reduce the overall accuracy.
The second approach, called "Multidimensional", does not use the target time series as input but rather other three correlated time series without gaps (soil temperature, air humidity, and photo-synthetically active radiation) in order to predict the values of stem moisture.
The advantage compared to the first case is the ability of this method to fill gaps of any length in a single step, without accumulating errors.
Although the second case is more successful when the gap lengths are arbitrarily long, the results in~\cite{song2020time} show that the first approach performs much better in filling sequences of a limited length.

Concerning the most effective "One-dimensional" approach, the authors in~\cite{song2020time} present two different LSTM-based forecasting networks and three different ways of combining them in order to improve the accuracy of the data filling.
The first network is called \textit{LSTM Forward} and it is the straightforward application of the LSTM architecture summarized in Table~\ref{table:Baseline_parameters}. 
This model is able to make predictions using only the values preceding the gap. 
The second network, which is called \textit{LSTM Reverse}, has the same architecture as the first one, but it is trained on the chronologically reversed version of the data. 
More concretely, it predicts the missing values using the data following the gap in a reverse order. 
Finally, the authors in~\cite{song2020time} present three different ways of combining the predictions of the \textit{LSTM Forward} and \textit{LSTM Reverse} models in order to exploit both the data preceding and following the gap they want to fill. 
All these three different methods of combining the forward and reversed models result in lower errors than the standalone \textit{LSTM Forward} and \textit{LSTM Reverse}. 
Among them, the best performing formula is the linear combination, called the \textit{decreasing weights} model in~\cite{song2020time}. 
More precisely, for every time step $i=0,\dots,n-1$, with $a_i$ being the forecasting of \textit{LSTM Forward} and $b_i$ the forecasting of the \textit{LSTM Reverse}, the resulting prediction is:
\begin{equation}
    z_i = (1 - c_i) a_i + c_i b_i, 
    \label{eq:decreasing weights}
\end{equation}
where $c_i = i/(n-1)$.
The linear combination~\eqref{eq:decreasing weights} of the \textit{LSTM Forward} and the \textit{LSTM Reverse} is thus the model from~\cite{song2020time} that we take as a baseline for the present work, and to which we refer as "Baseline model", or simply "Baseline", throughout the paper.

%
%
%

\section{Methodology} \label{methodology}

\subsection{Dataset description} \label{Dataset description}

This work is based on data that come from sensors located in monitored apartments in Bolzano, Italy, which were refurbished during the Sinfonia project. 
Sinfonia is an HORIZON 2020 five-year project, concluded in June 2020 and committed to the deployment of large-scale, integrated and scalable energy solutions in mid-sized European cities. 
Part of the activities carried out in Bolzano consisted of refurbishing social housing apartments, in order to pursue the maximum energy efficiency and a large use of renewable energy, improving also the comfort levels of the tenants.
The building groups that are available for analysis are located in four different sites of Bolzano: \textit{via Passeggiata dei Castani}, \textit{via Aslago}, \textit{via Similaun}, and \textit{via Brescia - via Cagliari}. 
In these locations, about 300 apartments were refurbished and about 120 of these are monitored using sensors, after obtaining the consent of tenants. 
More details about the locations of the buildings and their socioeconomic context can be found in~\cite{caballero2021tackling, dellavalle2018search}.

The monitoring system in the buildings is made of several sensors that track information about internal and external temperature, indoor relative humidity and CO$_2$ concentration, energy consumption for space heating, electricity, domestic hot water, energy production from different sources (photovoltaics, geothermal, solar thermal, boiler), and the opening and closure of windows in the apartments. 
In the context of this paper, the sensors that are used are those of internal temperature in the apartments (referred to as TEM sensor) and the sensor of external temperature (referred to as TEXT sensor). 

The final part of the Sinfonia project involved analysing the available data from the refurbished buildings in order to assess the results of the project and to ensure that it met its goals in terms of energy efficiency and tenants' comfort.
The analysis showed that data presented gaps distributed randomly, caused by the malfunctioning of sensors or by the interruption of data transmission. 
The frequency of these data gaps was not uniform across the different sites and across the sensors, meaning that some specific sites or sensor types were more affected than others, while other sensors had almost no missing values at all. 
Besides the missing values caused by the malfunctioning of the transmission system, failings of the sensors produced in some cases unreliable measurements (outliers) or constant values that had to be removed.
As a result, some sensors had missing values for longer than hours or even a whole day. This aspect made the usage of naive interpolation insufficient to fill the missing values due to their natural fluctuation.

With these premises, in this work we implement two algorithms that use part of the data that we have available in order to predict the missing values in the internal temperature sensor (TEM). More precisely, these algorithms use as input the TEM values preceding and following the data gap to be filled, together with all the TEXT values (that is, preceding, following, and during the TEM data gap).
We chose the TEM sensor as the target of our data filling since the internal temperature is a fundamental variable when assessing the comfort of the tenants. In addition, it shows a high correlation with the TEXT time series, which was available almost without gaps.
Concerning the data for the external temperature, only one TEXT sensor is available for the whole Sinfonia project in Bolzano, and it is located on the roof of the building in \textit{via Brescia - via Cagliari}.

Additionally, one of the requirements to train DL models for data filling is having a full data set for the target variable (the TEM sensor in our case). 
The reason is in the nature of DL models, that must be trained on uniform and reliable data in order to make accurate predictions once they have been trained.
Therefore, we decided to select apartments from one specific site of Bolzano that includes several apartments with full or nearly full data and fewer apartments with large data gaps on which the trained models could be applied. The data set of \textit{via Similaun} was the most appropriate data set for the target TEM sensor because it consists of 17 apartments with more than 96\% of data availability and 9 apartments with less than 90\% of data availability. 
The sensors in the \textit{via Similaun} buildings were placed at the end of 2019.
Consequently, we had reliable data for the purpose of this work from January 15th, 2020 until August 15th, 2021.
Concerning the other building sites, the one of \textit{via Aslago} was largely affected by technical errors on the sensors, while the site of \textit{via Brescia - via Cagliari} contains 25 apartments with about 90\% data availability and only 10 with more than 98\%. The data from the site of \textit{via Passeggiata dei Castani} is in a state comparable to that of \textit{via Similaun}. For computability reasons, we had to use only one of the two sites. 
Our final choice was \textit{via Similaun} due to the fact that \textit{via Brescia - via Cagliari}, where the only TEXT sensor is located, is closer to \textit{via Similaun} than \textit{via Passeggiata dei Castani}. Indeed, \textit{via Similaun} and \textit{via Brescia - via Cagliari} have similar urban contexts and external climatic conditions.
Thus, we obtain the data for the external temperature from a site that is 1.5 km far from the target site for the internal temperatures, that is the one of \textit{via Similaun}.
This is not an issue for the purposes of our work, since we are interested in the hourly and daily oscillations of the external temperature, which can be assumed to be sufficiently homogeneous between the two locations.

\subsection{Data preparation} \label{Data preparation}

In this section, the main initial decisions and the preparatory work on the data are explained. 
Those include:
\begin{itemize}
    \item the data analyses before the selection of the apartments and the time periods;
    \item the window size of the time series and the granularity of the sensor-incoming data;
    \item the separation of the data set into training-validation-test sets.
\end{itemize}

Firstly, the apartments from the buildings in \textit{via Similaun} had occurrences of malfunctioning TEM sensors that resulted in missing values, but much less frequently than in the other building sites. 
More precisely, the average length of the gaps per apartment over the whole period was between 0.58 and 0.78 days (average of all apartments 0.64 days), with standard deviation between 0.68 and 0.9 days (average of all apartments 0.74 days).
Those values had to be either removed or filled via interpolation before proceeding to the training of the models.
We decided to exclude those periods from the training and testing samples instead of interpolating them, since interpolation creates synthetic information in the input dataset. This removal is expected to have a very small impact on the characteristics and periodicity of the time series, while avoiding the elimination of the majority of the apartments.

In addition, defining the size of the data window is an important decision, and the possibilities are varied.
We decided to use 6 days of data for predicting the next 1 day, which sets the window size to 6-to-1 (or overall 7) days, even if this choice implies higher running complexity and requires more computational resources than using shorter periods (e.g. 20-to-4 hours). We decided to work with the 6-to-1 days window size for the following reasons:
\begin{itemize}
    \item After the analysis of the time series, we observed daily, weekly and seasonal periodicity in the internal temperature sequences. Leaving aside the seasonal range for computational reasons, choosing data sequences of 7 days allows capturing both daily and weekly patterns in the data. 
    \item We performed tests for different window sizes (20-to-4 hours, 1-to-1 days, 3-to-1 days, 6-to-1 days) and the models showed the best results on the 6-to-1 window frame. 
    Trying with longer frames is an interesting option, but it would require more computational resources and a longer time span as well, that were not available for the project.
    \item The majority of data gaps that we would like to actually fill are of shorter duration than one day.
\end{itemize}

Another consideration before proceeding to the actual development of the model is the data granularity. 
In the preliminary data analysis, we observed that the internal temperature in the apartments tends to have very small variations within a 15-minute interval, typically in the order of the accuracy range of the sensors installed.
Therefore, we decided to lower the granularity to a 15-minute interval using the average of the 5-minute values. 
In this way, we were able to increase the efficiency of the training of the models, without compromising the high granularity and having still 4 measurements per hour.

\begin{figure}[h!]
    \centering
    \includegraphics[width=\linewidth]{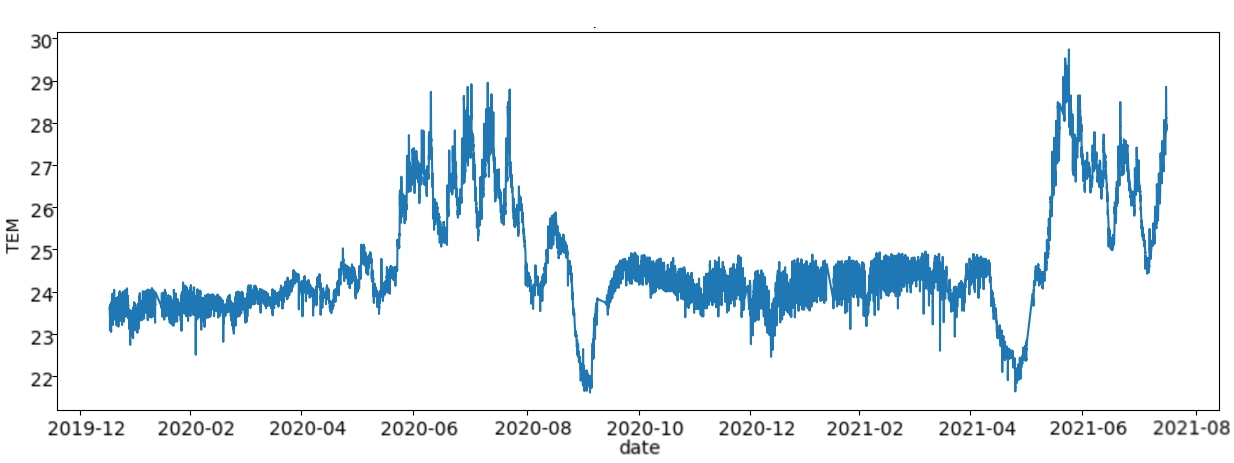}
    \caption{An example of TEM sensor data over the 19 months used for this work.}
    \label{fig:TEM_variations}
\end{figure}

The last preparation step consists of  splitting the time series samples into training, validation, and test sets.
Concerning the training set, we chose to use a whole year of data, from January 15th, 2020 until January 14th, 2021. 
The easiest and most naive option for splitting the remaining samples into validation and test sets would have been using the next 3.5 months as validation set and the remaining 3.5 months as test set.
Although this is normally chosen in cases of time series forecasting problems, we avoided it since the testing of the model would have been applied on data that belongs to only one season of the year (May 1st, 2021 - August 15th, 2021).
Indeed, as shown in Fig.~\ref{fig:TEM_variations}, during the summer months the internal temperature data show higher variations than in the winter period, in which the heating system is activated and the internal temperatures are more stable. 
The purpose of the testing phase, on the contrary, is to evaluate the models in all the different seasonal circumstances and without biases with respect to the training and validation sets.
Hence, we separated the 7 non-training months (January 15th, 2021 - August 15th, 2021) into:
\begin{itemize}
    \item a validation set that includes samples from the periods January 15th, 2021 - February 28th, 2021 and April 16th, 2021 - May 31st, 2021,
    \item a test set with samples from the periods March 1st, 2021 - April 15th, 2021 and June 1st, 2021 - August 15th, 2021.
\end{itemize}

In this way, the validation set includes a period when the space heating is activated (January 15th - February 28th) and another one in which it is deactivated (April 16th - May 31st). 
Similarly for the test set, the space heating is on for 1.5 months (March 1st - April 15th), but is off for the other 2.5 months (June 1st - August 15th). 

In the remaining part of this subsection, we describe how we define a uniform data structure starting from the separate TEM time series from multiple apartments and the TEXT sensor. This structure contains samples of fixed-length two-dimensional time series, and it is realized following also the guidelines in~\cite{brownlee2018deep}.

First, for each apartment the series of TEM sensor values is horizontally merged with the series of TEXT sensor values, creating two columns of series. 
During the training phase of the models, each column is treated as a distinct feature, justifying the notion of multivariate time series.
We point out that the data from all the selected apartments are used to train the models without considering the ID of the apartment they come from. 
In this way, we manage to process the available data sequences as if they all belonged to one archetype apartment. 
This is used both to have more training data and to avoid over-fitting to one single apartment, improving the ability of the models to generalize.
Finally, from each time series we extract sequences of values of fixed length, using a sliding window that moves one step (15 minutes) at a time for a total window size of 6-to-1 days. 
This procedure is repeated for all apartments and the sequences are simply added as samples in the general data structure. 

An additional step is to normalize the temperature values, both internal and external, in order to assist the training of the models. 
All values are scaled to a range between 0 and 1, while their previous range is approximately from -4 to 34. 
This task is accomplished by calculating the overall minimum and maximum values for each of the different kinds of temperatures and then applying the MinMax scaler from~\cite{scikit-learn}, as done in several studies~\cite{coutinho2018application, lara2021experimental, hong2020multivariate}. The examination of how different means could affect the accuracy of the models is out of the scope of this work.

After this pipeline is completed, the initial data is separated into three different time series sets (training, validation, test) in the way described above. 
However, their form is not yet completely appropriate for being used by the models. 
The reason is that, in order to forecast a specific day, the models receive as input 6 (past) days of TEM sensor series and 7 days (past and day of prediction) of TEXT sensor series.
Therefore, we separate each rolling split of one week into an input sample of 6 days of TEM sensor data and 7 days of TEXT sensor data, and an output sample of 1 day of TEM sensor data. 
Finally, after carrying out this procedure for all sets (training, validation, testing), the data samples are ready to be fed to the models for training.

%
%
\subsection{Models architecture}

In the following we outline the selection, implementation, and characteristics of the data-filling models that are developed in this work. 
For the sake of completeness, we specify that to create and evaluate the DL models of the present work, we used Python version 3.7.9 and the following packages:
\begin{itemize}
    \item TensorFlow~\cite{tensorflow2015-whitepaper} | version 2.4.1,
    \item Keras~\cite{chollet2015keras} | version 2.4.0,
    \item Scikit-learn~\cite{scikit-learn} | version 0.23.2.
\end{itemize}

%
%
\subsubsection{CNN-LSTM model} \label{CNN-LSTM model}

\begin{figure}[h!]
    \centering
    \includegraphics[width=\linewidth]{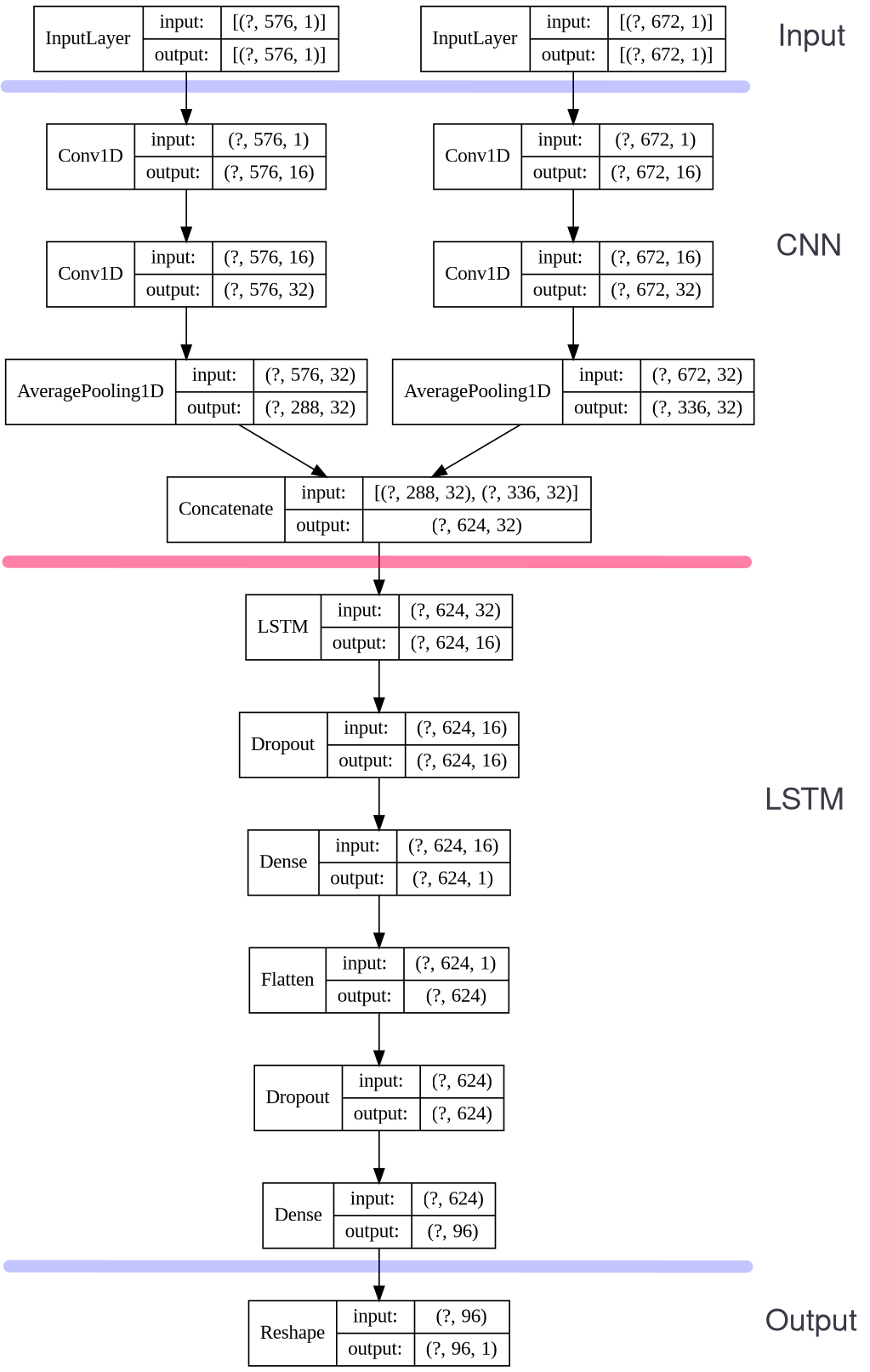}
    \caption{The CNN-LSTM networks.}
    \label{fig:CNN_LSTM_parallel_shapes}
\end{figure}

The first data filling model that we present in this work is based on the use of CNN and LSTM neural networks, and we refer to it as CNN-LSTM. 
This approach consists of two different DL networks, namely the CNN-LSTM-Onwards and the CNN-LSTM-Backwards, which are combined to produce the final predictions for the data filling.
The CNN-LSTM-Onwards and the CNN-LSTM-Backwards have the exact same neuronal structure, which is outlined in Fig.~\ref{fig:CNN_LSTM_parallel_shapes}, but differ in the data sequences they are trained with. 
Indeed, while the former model is trained on the time series with the usual chronological order, the latter is trained on the reversed form of the same time series. 
In other words, the first network learns to predict a sequence of 1 day using as input the past 6 days while the second network learns to predict 1 day using as input the following 6 days. 
Lastly, the two different predictions are combined in order to produce the final output of the CNN-LSTM data-filling model.
In this way, we exploit both the past and the future data in order to predict the target missing values. 

We will now describe the structure of the CNN-LSTM-Onwards and the CNN-LSTM-Backwards networks, which is identical and it is shown in Fig.~\ref{fig:CNN_LSTM_parallel_shapes}, and the way in which the model combines their predictions.
Firstly, the input sequences are time series samples of 6 days of TEM sensor and 7 days of TEXT sensor, with a 15-minute interval between the measurements.
The first sequence refers to the TEM sensor and has a shape\footnote{In this work, with \emph{shape} we refer to the dimensions of a tensor (data object used by the DL framework). 
The first number represents the time steps, while the second the number of features (either TEM, TEXT, or both).} of $(576, 1)$, since 6 days of TEM sensor values are used as input: \[6\ days\ \times \ 24\ hours\ \times \ 4\ values\ per\ hour\ =\]
\[576\ values.\ \ \ \ \ (TEM)\]
The second sequence, instead, regards the TEXT sensor and has a shape of $(672, 1)$, since TEXT data of 7 days is used as part of the input: \[7\ days\ \times \ 24\ hours\ \times \ 4\ values\ per\ hour\ =\]
\[672\ values.\ \ \ \ \ (TEXT)\]
As shown in Fig.~\ref{fig:CNN_LSTM_parallel_shapes}, which outlines the structure of the whole network, the above sequences are passed separately into two different Input Layers.

%
%
After this, two sub-networks of CNN architectures run in parallel and have the same type of layers but a different number of neurons. This part, which is the first core of the network, has the scope of extracting key information from the inputs.
The left part in Fig.~\ref{fig:CNN_LSTM_parallel_shapes} receives data from the TEM sensor, while the right one from the TEXT sensor. In both, the first layer is a convolutional layer of one dimension (named Conv1D) with 16 neurons, 3 kernels, and the Rectified Linear Unit (ReLU) activation function.
Afterwards, a second convolutional layer of the same type is applied with 32 neurons this time, but the same activation function and number of kernels.

Concerning the selection of the hyper-parameters of the Conv1D, we specified the length of the 1D convolution filter (kernel size) and the number of output filters in the convolution.
Both were set manually after numerous tests by inspecting the models accuracy. 
In addition, we ran tests using different activation functions (sigmoid, hyperbolic tangent, softmax, and ReLU) and even no activation at all.
The lowest errors were achieved using the ReLU activation, which is indeed one of the most used in DL architectures. 
As described for instance in~\cite[p.~3]{rai2020cnn}, ReLU "effectively handles the gradient vanishing as well as gradient exploding problems".

As a next step, the size of the outputs of the CNN layers is reduced by half after applying an Average Pooling layer in both parallel pipelines. 
The usage of Max Pooling was also tested as an alternative, but the errors of the results were lower with Average Pooling.
Then, in the last step of the CNN part of the model, the outputs of the two sub-networks are concatenated on the same number of neurons (32) in order to save all the information extracted from the TEM and TEXT series. As shown in Fig.~\ref{fig:CNN_LSTM_parallel_shapes}, the shape of the layer resulting from the CNN part is $(624, 32)$.

%
%
In the second half of the network architecture, a single LSTM layer is applied, as shown in Fig.~\ref{fig:CNN_LSTM_parallel_shapes}. 
The LSTM layer receives data with shape $(624, 32)$, and it is applied using 16 neurons. 
Also, it is run enabling the return of the hidden units, instead of simply returning as output the last (output) layer of the LSTM structure. 
This results in having a 2-dimensional array with shape $(624, 16)$, instead of $(1, 16)$, as the output of the LSTM. 
In that way, more information is transferred to the following layers as the structure approaches the final prediction. 

After the LSTM layer, two Fully Connected (Dense) layers are designed to reduce the output size matching the desired shape of the final prediction. 
To avoid the over-fitting of the model, we add Dropout layers after the LSTM and between the Dense layers for dropping 10\% of the units before moving to the next layer. 
As described in~\cite[p.~88]{krizhevsky2017imagenet}, in this way the model is "\dots forced to learn more robust features that are useful in conjunction with many different random subsets of the other neurons".

The possibility of inserting Dropout layers also between Pooling, CNN, or LSTM layers was explored, as done in~\cite{rai2020cnn, hong2020multivariate}, but the errors of the predictions were significantly higher. 
We had the lowest errors when using Dropout after the LSTM layers and between the Dense layers. This is a standard procedure in the recent literature, followed for instance in~\cite{fu2019multi, le2019improving}.
Finally, the output shape is $(96, 1)$, which is the number of timestamps covering our target prediction of a whole day.
Table~\ref{table:cnn_lstm_parameters} below shows the number of neurons parameters in each layer. 

\begin{table}[h!]
    \centering
    \caption{The layers and trainable parameters of the CNN-LSTM networks.}
    \label{table:cnn_lstm_parameters}
    \begin{tabular}{c c c} 
         Layers & Neurons & Parameters \\ [1ex] 
         \hline\hline \\
         Conv1D & 16 & 32 \\ 
         Conv1D & 16 & 32 \\
         Conv1D & 32 & 544 \\
         Conv1D & 32 & 544 \\
         LSTM & 16 & 3,136 \\
         Dense & 1 & 17 \\
         Dense & 96 & 60,000 \\ [1ex]
         \hline\hline \\
         Total parameters & & 64,305 \\ [1ex]
         \hline\hline \\
    \end{tabular}
\end{table}

%
Up to this point, we described the common structure of the two networks in the CNN-LSTM model. Despite they have the same structure, the two networks are trained separately and without any interaction: the CNN-LSTM-Onwards on time series with the usual chronological order, and the CNN-LSTM-Backwards with the reversed time series, as explained above.
After their training is completed, they are used for forecasting forward and backward in time accordingly.
Moreover, the last key step for obtaining the CNN-LSTM model predictions is to combine the two different predicted sequences coming from CNN-LSTM-Onwards and CNN-LSTM-Backwards into one single sequence of predicted data. 
The motivation of combining the forward and backward predictions lies in the observations of the results of each model separately and it was already part of the approach in~\cite{song2020time}. 
Indeed, the forecasting of the forward network shows lower errors in the first part of the predicted sequence, while the predictions for the last time steps are often unable to follow the real trend of the series. 
Similarly, but in a symmetric way, the backward network is much more accurate in the last time steps of the data gap. 
Hence, we combine the outputs from the two networks using the sigmoid function.
Specifically, we create a 96-points sequence from -6 to 6, with a constant step. Then, we transform this sequence using the sigmoid function. 
The final output of the CNN-LSTM model is the result of the following combination
\begin{equation} \label{sigmoid_pred}
    P_i = (1-s_i) \cdot f_i + s_i \cdot b_i,
\end{equation}
where for each time step $i$, $s_i$ is the $i_{th}$ weight from the Sigmoid transformation, $f_i$ is the forward prediction, $b_i$ the backward prediction, and $P_i$ the final prediction.

\begin{figure}[h!]
\centering
\begin{subfigure}{\linewidth}
  \centering
  \includegraphics[width=1\linewidth]{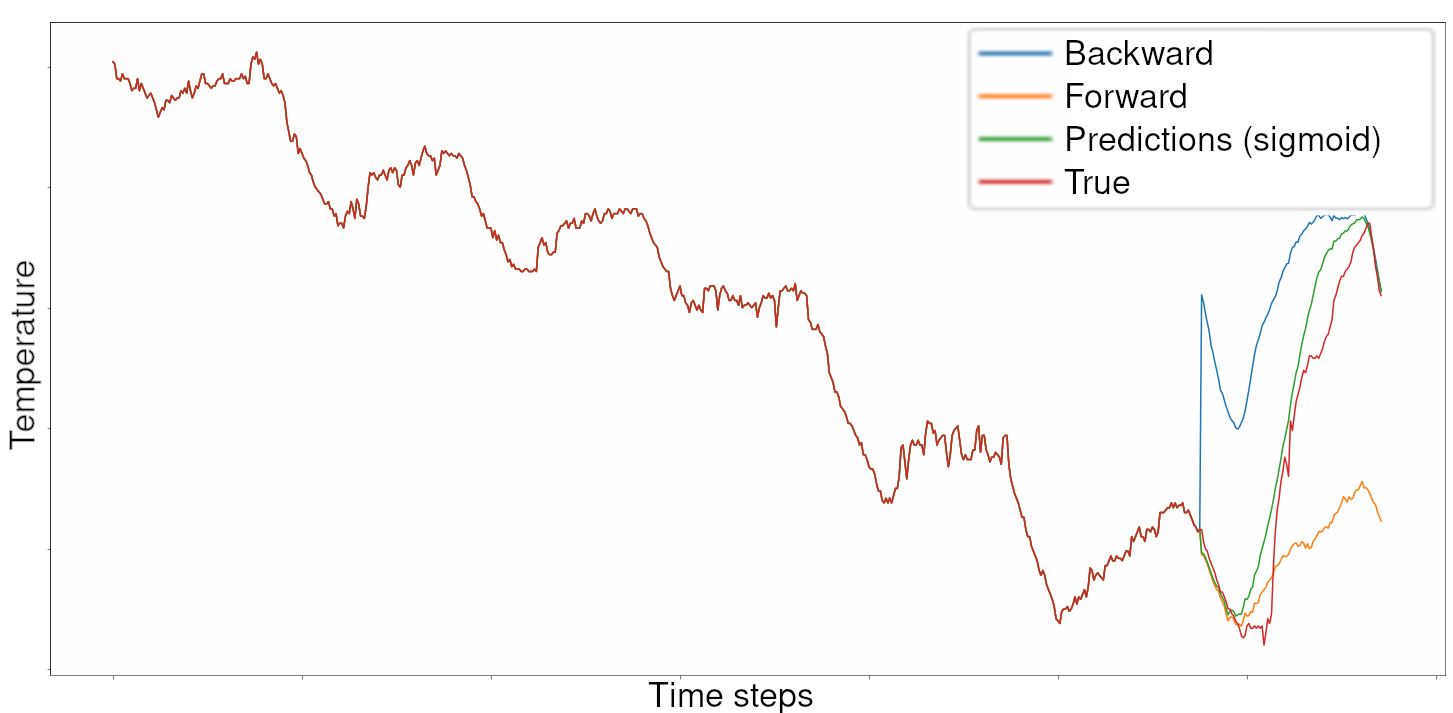}
  \caption{Entire plot.}
  \label{fig:sigmoid_testcase}
\end{subfigure}
\begin{subfigure}{\linewidth}
  \centering
  \includegraphics[width=0.4\linewidth]{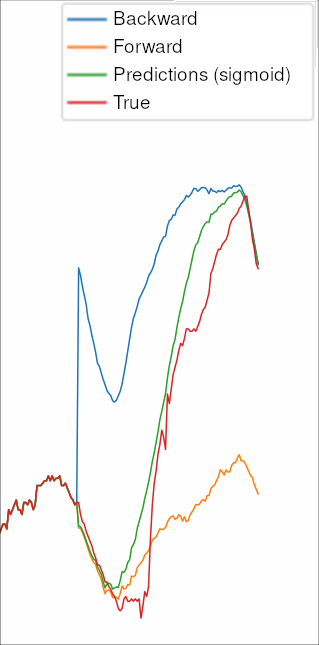}
  \caption{Cropped plot zooming on the predictions.}
  \label{fig:sigmoid_testcase_Zoomed}
\end{subfigure}
\caption{A 6-days time series and the predictions for the following day using the CNN-LSTM model.}
\end{figure}

Using the sigmoid function allows to weight the predictions of the forward and backward models in a time-dependent way. 
The idea is that for each timestamp the combined model gives more weight to the prediction that is more likely to be accurate.
In Fig.~\ref{fig:sigmoid_testcase}, we provide an example of the predictions made by the CNN-LSTM model on a sample from our test data set, in order to understand the contribution of the two separate networks and of their combination via formula~\eqref{sigmoid_pred}. 
In Fig.~\ref{fig:sigmoid_testcase}, the past 6 days and the real values of the target day are plotted (in red) along with the predictions from the CNN-LSTM-Backwards (in blue), the CNN-LSTM-Onwards (in yellow), and their sigmoid combination (in green) which is the final output of the CNN-LSTM model. 
The 6 days following the target day are also provided to the model as input, even if they are not plotted here for visibility purposes.
Focusing on the predictions made for the last day displayed in Fig.~\ref{fig:sigmoid_testcase_Zoomed}, the three different sequences show the behavior described earlier. 
In particular, it is important to highlight that in this specific examples the true values (in red) show a trend reversal during the prediction day with respect to the previous six days. This is only partially captured by the forward prediction, that underestimates the temperature rise in the second part of the gap. This is balanced by the backward prediction, that instead  is closer to the real values in the last part of the interval.
As a result, their sigmoid combination (in green) achieves the best result, being the closest to the real values (in red) throughout the prediction day. This example clearly shows the importance of considering also the data after the data gap as inputs of a data filling problem for time series.
For a complete assessment of the model results we refer to Section~\ref{Results and discussion}.

%
%
\smallskip
\subsubsection{CNN-BiLSTM model} \label{CNN-BiLSTM model}

The second data filling model that we present in this work is based on the use of CNN and BiLSTM neural networks, and we refer to it as "CNN-BiLSTM". 
In contrast with the CNN-LSTM model outlined in the previous section, this approach uses only one DL network and obtains the final predictions directly as the outputs of the network itself.
Despite of this structural difference, the length of the gaps that are filled by the CNN-BiLSTM model remains the same as the CNN-LSTM model, that is, spanning 1 day of measurements.

It is important to stress that the model uses as input
\begin{itemize}
    \item 12 days of TEM: the past 6 days and the following 6 days of the TEM sensor time series;
    \item 13 days of TEXT: the past 6 days, the day corresponding to the prediction, and the following 6 days of the TEXT sensor time series.
\end{itemize}

%
In contrast to our first model that uses forward and backwards inputs, the input data for the CNN-BiLSTM model consists of time series in the usual chronological order.
The sequences are initially prepared as described in Section~\ref{Data preparation}, where the time series are transformed into samples with 7 days of data (6 days as input - 1 day for prediction). 
In addition, to create windows of 13 days, we concatenate the following 6 days of data at the end of each 7-days sequence. 

\begin{figure}[h!]
    \centering
    \includegraphics[width=\linewidth]{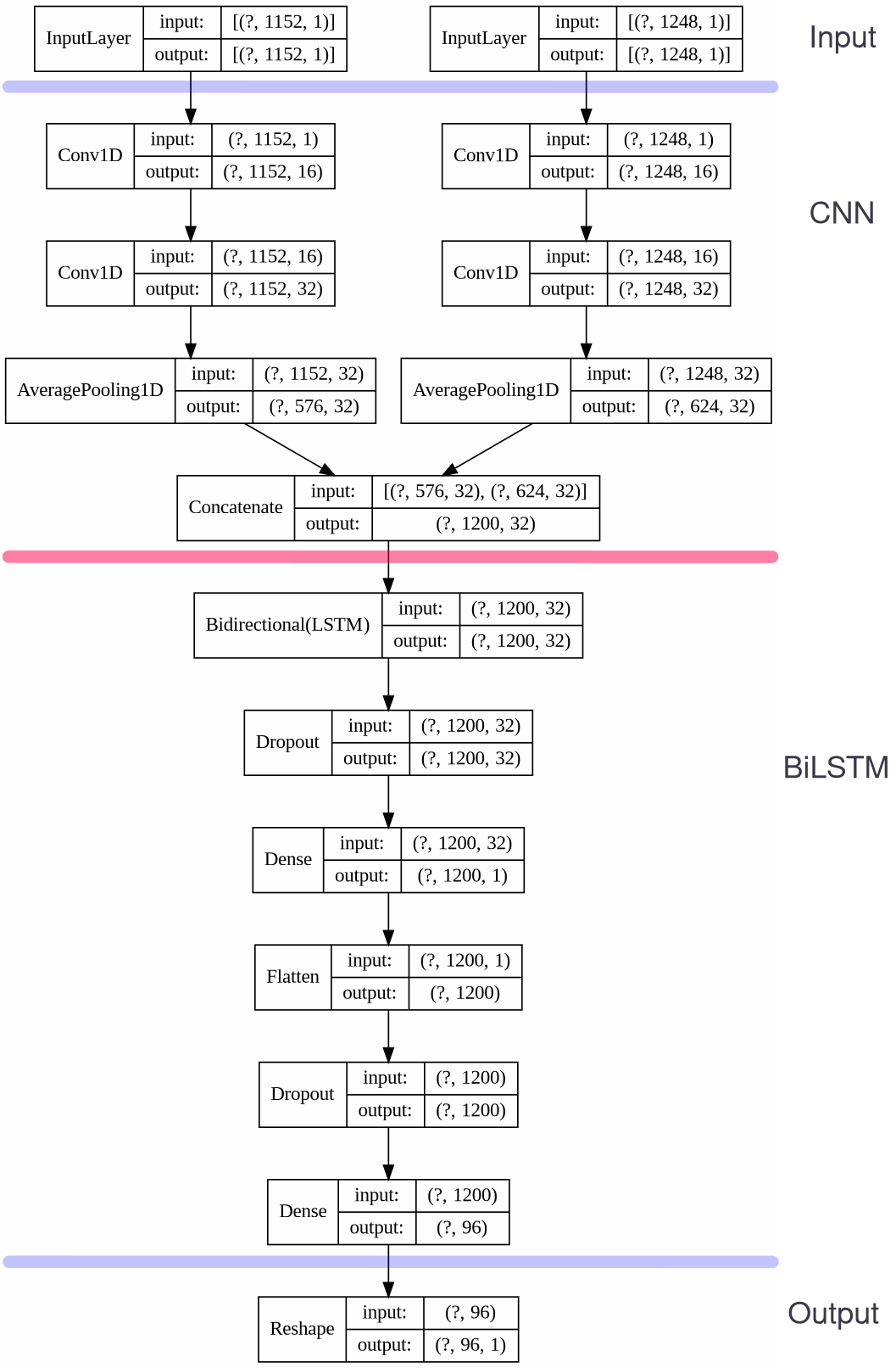}
    \caption{The CNN-BiLSTM model structure.}
    \label{fig:CNN_BiLSTM_parallel_shapes}
\end{figure}

The TEM sensor sequence fed into the CNN-BiLSTM model has a shape of $(1152, 1)$, since 12 days of TEM sensor values are used as input: \[12\ days\ \times \ 24\ hours\ \times \ 4\ values\ per\ hour\ =\]
\[1152\ values.\ \ \ \ \ (TEM)\]
The second input sequence, containing data from the TEXT sensor, has a shape of $(1248, 1)$, since TEXT data of 13 days is used as part of the input: \[13\ days\ \times \ 24\ hours\ \times \ 4\ values\ per\ hour\ =\]
\[1248\ values.\ \ \ \ \ (TEXT)\]
Consequently, the above sequences are passed separately to two different input layers. 
The shapes of the layers can be seen in Fig.~\ref{fig:CNN_BiLSTM_parallel_shapes}, which represents the structure of the whole model.

%
%
The first core part of the CNN-BiLSTM model is a convolutional architecture, identical to the one of the CNN-LSTM networks shown in Fig.~\ref{fig:CNN_LSTM_parallel_shapes}.
We decided to maintain the same CNN-part in all networks in order to focus our investigation on the impacts of using different LSTM architectures.
Concerning the CNN part of the model, the only difference here with respect to the CNN-LSTM network is in the shapes of the tensors, which is a straightforward consequence of having input data of different length.
For the details about the CNN architecture we thus refer to Section~\ref{CNN-LSTM model}. 
As shown in Fig.~\ref{fig:CNN_BiLSTM_parallel_shapes}, the shape of the resulting layer is $(1200, 32)$.

%
%
The second main component of the model architecture is the BiLSTM network, starting with a single Bidirectional LSTM layer.
The input to the BiLSTM layer has a shape of $(1200, 32)$ and its output consists of 16 neurons. 
As it is done in the case of the LSTM layers in the CNN-LSTM networks described above, we run it with the return of the hidden units enabled.
Therefore, the output of the BiLSTM layer is of shape $(1200, 32)$. 
The main difference now is that the BiLSTM architecture consists of a combination of a Bidirectional layer and an LSTM layer. The Bidirectional layer connects two hidden layers --- the two LSTMs running in the opposite directions --- to the same output. 
In this way, the output layer gets information from past and future states simultaneously, as explained in detail in Section~\ref{BiLSTM}.

The final step is to use two Fully Connected (Dense) layers combined with Dropout layers for creating the final output while also avoiding the over-fitting of the model.
The architecture of this part is identical to the one of the CNN-LSTM model and motivated accordingly. 
At the end, the final output has the target shape of the 1-day prediction, which is $(96, 1)$.
As for the first model, we outline the number of neurons and parameters per layer in a table, that for the CNN-BiLSTM model is Table~\ref{table:cnn_bilstm_parameters} below.  

\begin{table}[h!]
    \centering
    \caption{The layers and trainable parameters of the CNN-BiLSTM model.}
    \label{table:cnn_bilstm_parameters}
    \begin{tabular}{c c c} 
         Layers & Neurons & Parameters \\ [1ex] 
         \hline\hline \\
         Conv1D & 16 & 32 \\ 
         Conv1D & 16 & 32 \\
         Conv1D & 32 & 544 \\
         Conv1D & 32 & 544 \\
         BiLSTM & 32 & 6,272 \\
         Dense & 1 & 33 \\
         Dense & 96 & 115,296 \\ [1ex]
         \hline\hline \\
         Total parameters & & 122,753 \\ [1ex]
         \hline\hline \\
    \end{tabular}
\end{table}

%
%
%

\section{Discussion of the results} \label{Results and discussion}

In this section, we outline the results achieved by the models presented in the previous chapter (CNN-LSTM and CNN-BiLSTM), compared to the Baseline model from~\cite{song2020time} described in Section~\ref{Baseline model}.
We point out that the training, validation, and test time series are the same for all three models and we refer to Section~\ref{Data preparation} for the details of the data preparation.
Besides this, we clarify that we ran all three data-filling models under the same conditions of hyper-parameters.

First, we set the same maximum number of epochs to 100 for all three models.
Additionally, an early stopping (callback) of 20 epochs is applied, allowing the training to continue for maximum 20 epochs even if it does not improve its validation loss.
These numbers are selected as a result of several tests from which we recognised that:
\begin{itemize}
    \item 20 epochs as early stopping is a balanced trade-off to allow the models to "have patience" until the possibility of a late improvement but also restricting them from running too long when there is no improvement,
    \item 100 epochs is a safe upper limit for the maximum epochs to be run since in all tests we made, due to the early stopping, the models never surpassed 85 epochs.
\end{itemize}

\begin{figure}[h!]
    \centering
    \begin{subfigure}{\linewidth}
        \centering
        \includegraphics[width=\linewidth]{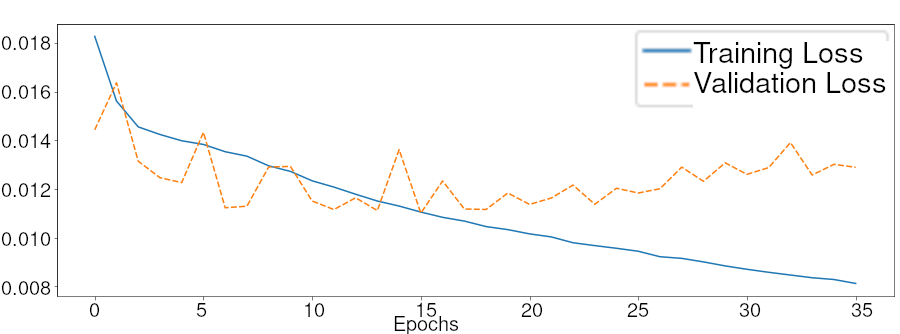}
    \end{subfigure}
            
    \begin{subfigure}{\linewidth}
        \centering
        \includegraphics[width=\linewidth]{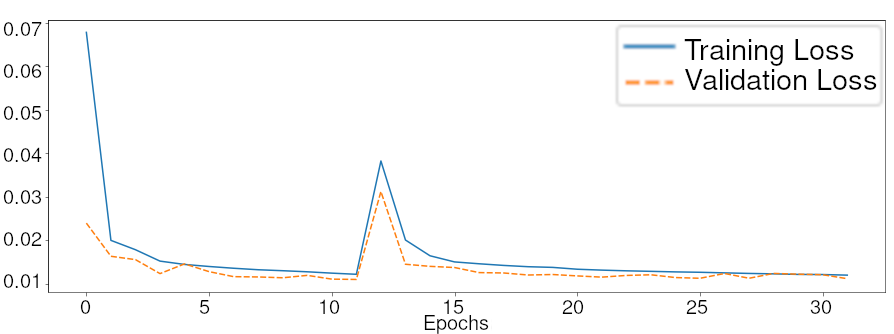}
    \end{subfigure}
    
    \caption{Training and validation losses of the Baseline model: LSTM-Forward above, LSTM-Reverse below.}
    \label{fig:train_val_losses_b}
\end{figure}

\begin{figure}[h!]
    \centering
    \begin{subfigure}{\linewidth}
        \centering
        \includegraphics[width=\linewidth]{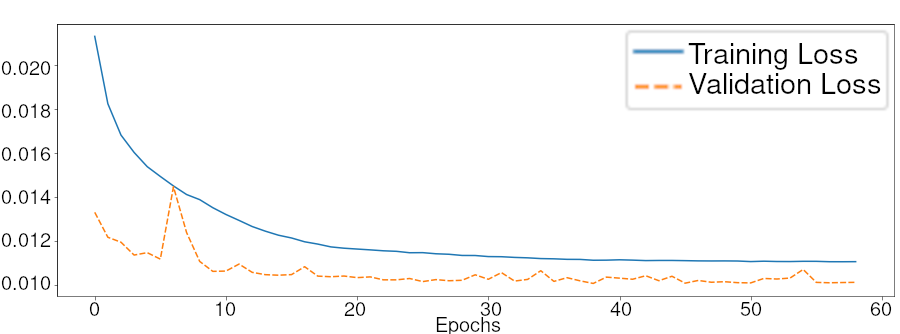}
    \end{subfigure}
    
    \begin{subfigure}{\linewidth}
        \centering
        \includegraphics[width=\linewidth]{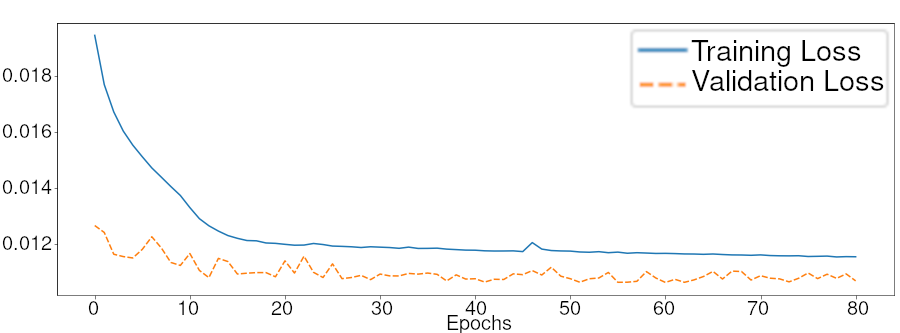}
    \end{subfigure}
    \caption{Training and validation losses of the CNN-LSTM-Onwards (above) and the CNN-LSTM-Backwards (below).}
    \label{fig:train_val_losses_L}
\end{figure}

\begin{figure}[h!]
    \centering
    \includegraphics[width=\linewidth]{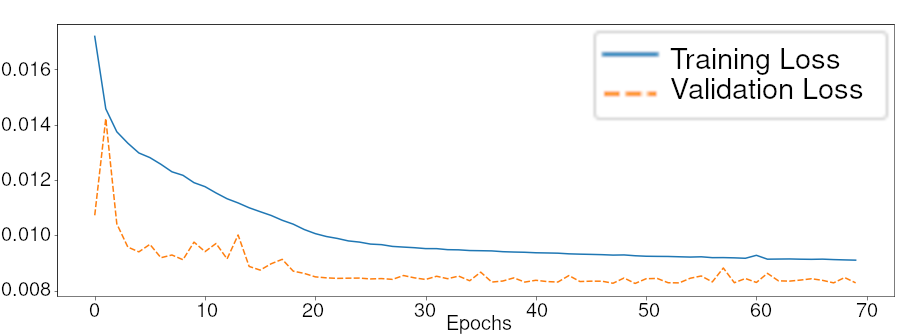}
    \caption{Training and validation losses of CNN-BiLSTM network.}
    \label{fig:train_val_losses_Bi}
\end{figure}

All three models show a similar behavior during the training: they achieve a very low training error before reaching 30 epochs and keep improving it slightly afterwards. 
On the contrary, the validation errors oscillate periodically since the first 20 epochs at latest, showing an expected trend of first over-fitting and then escaping from it.
In Fig.~\ref{fig:train_val_losses_b}, \ref{fig:train_val_losses_L}, and~\ref{fig:train_val_losses_Bi}, we show the training and validation losses during the training of all the networks composing the three models. 

Another hyper-parameter that we set equal for all the models is the batch size, that is the number of samples that are propagated through the network training at each iteration. 
We set the batch size to 512 in order to reduce the duration of the training for each model, since the number of samples was more than half a million.

Concerning the loss function, we point out that we use the same one (the Mean Absolute Error, see \eqref{eq:MAEeq}) for the training of all three models. This allows us to fairly compare how fast the models are trained and how they behave with respect to over-fitting. 

\subsection{Performance metrics} \label{Evaluation metrics}

In the following we briefly describe the evaluation metrics we used during the training and testing of our models. Apart from one custom metric we define, all the other functions are the standard ones used in the literature for similar data-filling problems --- see for instance~\cite{livieris2020cnn, rai2020cnn, kim2019predicting, cai2019day, lara2021experimental, cheng2019multi}. 
%
%
Three classical error metrics that we use are the Mean Absolute Error (MAE), the Mean Squared Error (MSE), and the Mean Absolute Percentage Error (MAPE). To define them, for every array extracted from a time series we take $n$ as the total number of time steps, $i=1,\dots, n$ as the pointer of each time step, $P_i$ as the predicted values, and $T_i$ as the true values. Then, the metrics are defined according to the formulas:
\begin{equation}
        MSE = \frac{1}{n} \sum_{i=1}^{n} (P_{i} - T_{i})^2,
        \label{eq:MSEeq}
\end{equation}
\begin{equation}
        MAE = \frac{1}{n} \sum_{i=1}^{n}\left | P_{i} - T_{i} \right |,
        \label{eq:MAEeq}
\end{equation}
\begin{equation}
        MAPE = \frac{100}{n} \sum_{i=1}^{n} \left | \frac{P_{i} - T_{i}}{T_{i}} \right |.
        \label{eq:MAPEeq}
\end{equation}

Using more than one metric to assess the results of data-filling models is indeed a standard procedure, which is mainly due to the fact that different metrics allows to evaluate different characteristics of the proposed models.
First, the MSE and MAE are the most commonly used metrics in a large variety of statistical and machine learning problems, including time series forecasting and data-filling. The difference among them lies in applying the power 2 (MSE) or the absolute value (MAE) to the difference between true and predicted values --- see~\eqref{eq:MSEeq}-\eqref{eq:MAEeq}. This difference normally results in the MSE being more affected by outliers than the MAE, due to the fact that the power 2 penalises large errors more than the absolute value does. Nevertheless, it is important to stress that when the data ranges between 0 and 1, as it is in our case during the training phase, the roles are reversed and the errors computed with MSE will be lower than the ones computed with MAE.
Next, the MAPE allows to make a significant comparison between models developed for different data filling problems.
Indeed, being a percentage error, it eliminates the biases given by the different measurement units and range of values.
However, a downside of MAPE --- see \eqref{eq:MAPEeq} --- is that its values may diverge when the true values are close to 0.
We point out that, indeed, we use the MAPE only in the testing phase, that is when the internal temperature values are in their original form, that in our dataset is always above 10°C.

During the training phase of the models, we used the MAE as the unique loss function for the fitting of the models and the MSE as an additional metric to assess how fast the models improve their performances. 
Indeed, since our training data is scaled to the range $(0,1)$, MAE penalises large errors more than the MSE and thus is more indicated to be used as loss function.
During the test phase, instead, we used all the three metrics defined above together with the $R^2$ and the MSTDR that we are going to define.

%
The $R^2$ Score, or Coefficient of Determination, is calculated for each of the 96 predicted values by the models (1 day divided into 15-minutes intervals). For every $i=1,\dots,96$, we define 
\begin{equation}
        R_i^2 = 1 - \frac{\sum_{j=1}^{N}(P_{j,i}-T_{j,i})^2}{\sum_{j=1}^{N}(\overline{T}_i-T_{j,i})^2},
        \label{eq:R2eq_point}
\end{equation}
where $j=1,\dots,N$ is the pointer of each 1-day sample predicted by the model, $P_{j,i}$ and $T_{j,i}$ are respectively the predicted and true $i$-th value in the $j$-th sample, and $\overline{T}_i$ is the average of the $i$-th true value over all the samples: $\overline{T}_i = \frac1N\sum_{j=1}^{N}T_{j,i}$.

To have a scalar version of this metric, we define its average over $i=1,\dots,96$ as
\begin{equation}
        \overline{R^2} = \frac{1}{96}\sum_{i=1}^{96}R_i^2.
        \label{eq:R2eq_ave}
\end{equation}
The $R^2_i$ is, at each point $i$, a statistical measure of how well that point of the sequence is predicted with respect to the average of its true values. 
It can assume negative values and its maximum value (corresponding to perfect predictions) is~1. 
Several studies --- see for instance~\cite{chicco2021coefficient} --- point out a big advantage of $R^2$ against other evaluation metrics in regression analysis.
Indeed, MSE, MAE, and MAPE values do not reveal the regression's quality with respect to the distribution of the true values. 
Instead, the $R^2$ can distinguish two apparently equally (in terms of MAE for instance) accurate models by quantifying how much better (or worse) are the predictions given by each model with respect to the simple average of the true values. That is, a model always predicting the mean of the real values, which is very far from what we aim at, might reach a good MSE or MAE, but its $R^2$ score will be 0.

%
Finally, in this paper we use also a custom metric, the Mean Standard Deviation Ratio (MSTDR), that calculates the mean of the ratios between the Standard Deviation (STD) of the predicted values and the STD of the true values. For every 1-day sample $j=1,\dots,N$ predicted by the model, we define the Standard Deviation Ratio (STDR) as
\begin{equation}
    STDR_j = \frac{STD(P_j)}{STD(T_j)},
    \label{eq:STDReq}
\end{equation}
where $T_j$ is the vector of 96 true values of sample $j$, and $P_j$ the predicted vector for the same sample.
Then, the MSTDR is defined as the average of the STD ratios over the samples, that is
\begin{equation}
    MSTDR = \frac{1}{N} \sum_{j=1}^{N} SDTR_j,
    \label{eq:MSTDReq}
\end{equation}
where $N$ is the total number of samples over which the average is taken.

The MSTDR expresses the similarity of the STD of the predicted values with the STD of the real values. 
Being a ratio of two positive quantities, the MSDTR ranges from 0 to infinity, with 1 denoting identical STDs.
The important difference with the $R^2$ metric is that in~\eqref{eq:R2eq_point} the average is taken over the $N$ samples in the dataset, while in~\eqref{eq:STDReq} the standard deviation is computed over the 96 values that compose a single sample. In this way, $STDR_j$ is a measure of how well the model represents the variations in the original $j$-th sequence. The MSTDR is then simply the average over all the samples of the $STDR_j$, that summarizes how well the model reproduces the oscillating nature of the true values.

\subsection{Errors and models comparison} \label{Errors comparison}
We outline here the results of the testing phase of our models, that we divided in two parts. We recall that all three models have been trained with data coming from 17 apartments located in \textit{via Similaun} (SIM) for the period from January 15th, 2020 until January 14th, 2021. 
Thus, we first test the models on data coming from the same apartments used for training, but for a different time period, as discussed in~\ref{Data preparation}. Then, we test the models also on data coming from apartments located in a different site, the one of \textit{Via Brescia - Via Cagliari} (BRE-CAG). This additional test is made to assess the ability of the models to perform on data coming from totally new sources, given the correlation between data coming from the same apartments in different periods. 

We begin by showing in Table~\ref{table:errors} the results of the models on SIM apartments for the test period.
In the upper half of the table, with the error metrics MSE, MAE, and MAPE, lower values mean that the model is performing better.
On the contrary, in the lower half of the table containing the score metrics $\overline{R^2}$ and MSDTR, the values are better as they get closer to~1.
The first comparison of the models predictions shows that all three models are sufficiently accurate. 
In particular, all the error metrics appoint the CNN-BiLSTM as the best performing model (values in \textbf{bold)}, while the other two perform similarly, with the CNN-LSTM having mildly higher errors than the Baseline.
\begin{table}[h!]
    \centering
    \caption{The results of the three models on the test set --- SIM apartments.}
    \label{table:errors}
    \begin{tabularx}{\columnwidth}{c | c c c } 
          & Baseline & CNN-LSTM & CNN-BiLSTM \\ [1ex]
         \hline\hline \\
         MSE & 0.124 & 0.138 & \textbf{0.068} \\
         MAE & 0.24 & 0.257 & \textbf{0.176} \\
         MAPE & 0.958 & 1.023 & \textbf{0.709} \\ [1ex]
         \hline \\
         $\overline{R^2}$ & 0.965 & 0.961 & \textbf{0.981} \\
         MSDTR & 0.752 & 0.919 & \textbf{0.938} \\ [1ex]
         \hline\hline \\
    \end{tabularx}
\end{table}
Besides, the $\overline{R^2}$ values are very similar among the three models, with the CNN-BiLSTM predictions showing a slightly better correlation to the real values than the other two. 
Concerning the MSDTR, our two models achieve a very high similarity of the predictions STDs with the STDs of the real values (above 90\%), while the Baseline model shows lower similarity (around 75\%). 
This fact suggests that the Baseline model is capturing less variability in the data than our models do.
That is, even though the CNN-LTSM shows the highest errors among the three, its predictions fluctuates more realistically than those of the Baseline. 
Nevertheless, the main outcome of Table~\ref{table:errors} is that the CNN-BiLSTM shows both the lowest errors and the most realistic approximations. 

\begin{figure}[h!]
    \centering
    \includegraphics[width=0.8\linewidth]{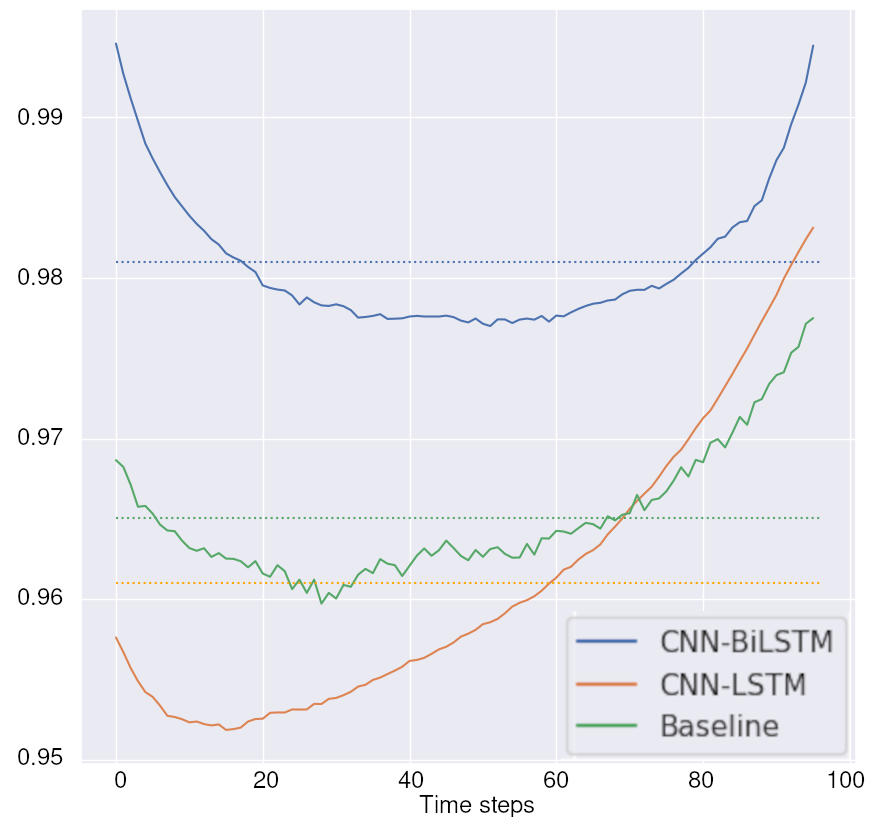}
    \caption{$R^2$ and their means ($\overline{R^2}$, dotted lines) for all models on the test set --- SIM apartments. 
    }
    \label{fig:r2s_distrib}
\end{figure}

As mentioned in Section~\ref{Evaluation metrics}, the $R^2$ is a significant measure of correlation between each time step's predictions and the real values. 
Thus, we inspect also its distribution across the whole window of predicted data instead of considering only its final mean $\overline{R^2}$. 
In Fig.~\ref{fig:r2s_distrib}, we show the $R^2$ values for all three models for each time step of the 1-day target prediction, together with their mean $\overline{R^2}$ plotted as dotted lines. 
From Fig.~\ref{fig:r2s_distrib}, we observe that:
\begin{itemize}
    \item The CNN-BiLSTM has the best performing approximation of the time series, both overall and of every individual time step.
    We can also observe how the performances are better in the first and last part of the 1-day forecast, due to the more proximity to the available data, and how symmetric those performances are, suggesting that the model exploits both past and future data in a similar way.
    \item Both the CNN-LSTM and the Baseline model show a higher $R^2$ score in the last part of the window with respect to the initial one.
    This suggests that the backward networks are performing better than the forward ones and that, arguably, future sequences (6 following days) are more effective to the data-filling result than the past sequences (6 preceding days).
    \item Although the Baseline model shows better $R^2$ values from the beginning until the 70th time step, the CNN-LSTM model overtakes it after that point. 
    Besides the differences in the networks, this could also be due the selection of the sigmoid formula~\eqref{sigmoid_pred} over the linear combination~\eqref{eq:decreasing weights} of the forward and backward predictions. 
    Indeed, the sigmoid combination strengthens the very first and very early predictions more than the linear combination does.
\end{itemize}

As mentioned above, in the testing phase we also aim to evaluate our models on data coming from apartments that has not been accessed by the models and are completely unrelated to the data used for training. 
Particularly, we outline in the following the results of the models on data from apartments belonging to the \textit{via  Brescia - via Cagliari} (BRE-CAG) site, while the training phase was done on apartments located in \textit{via Similaun} (SIM). The period in which we ran the tests are the same for both groups of apartments and are discussed in Section~\ref{Data preparation}. 
Therefore, Table~\ref{table:errors_unseen} shows the errors and scores of all three models, similarly to Table~\ref{table:errors} above, but this time for data coming from BRE-CAG apartments. 

\begin{table}[h!]
    \centering
    \caption{The results of the three models on the test set --- BRE-CAG apartments.}
    \label{table:errors_unseen}
    \begin{tabularx}{\columnwidth}{ c | c c c}
          & Baseline & CNN-LSTM & CNN-BiLSTM \\ [1ex] 
         \hline\hline \\
         MSE & 0.606 & 0.366 & \textbf{0.273} \\
         MAE & 0.371 & 0.359 & \textbf{0.272} \\
         MAPE & 1.683 & 1.518 & \textbf{1.197} \\ [1ex]
         \hline \\
         $\overline{R^2}$ & 0.886 & 0.931 & \textbf{0.948} \\
         MSDTR & 0.666 & \textbf{0.834} & 0.83 \\ [1ex]
         \hline\hline \\
    \end{tabularx}
\end{table}

Since the BRE-CAG apartments are "unseen" by the models in their training phase, the higher errors in Table~\ref{table:errors_unseen} with respect to Table~\ref{table:errors} are in line with the expectations.
As in the case of SIM apartments, the CNN-BiLSTM shows the best results among the three models.
However, it is interesting to notice how the CNN-LSTM model worsens its performance in the error metrics (MSE, MAE, MAPE) much less than the Baseline does.
This suggests a higher ability of the CNN-LSTM model in generalising its predictions compared to the Baseline.
Indeed, this is confirmed by the $\overline{R^2}$ and the MSDTR, with the CNN-LSTM and the BiLSTM clearly outperforming the Baseline.
We must point out that for the first time the CNN-BiLSTM has a slightly worse result than the CNN-LSTM, with the latter having the highest MSDTR, even though the two results are very similar.
In general, these results show that our models perform understandably worse on totally new apartments, but the performances are still good, and sensibly better than the Baseline model. This indicates a good ability of our models in generalizing to new data.

\begin{figure}[h!]
    \centering
    \includegraphics[width=0.8\linewidth]{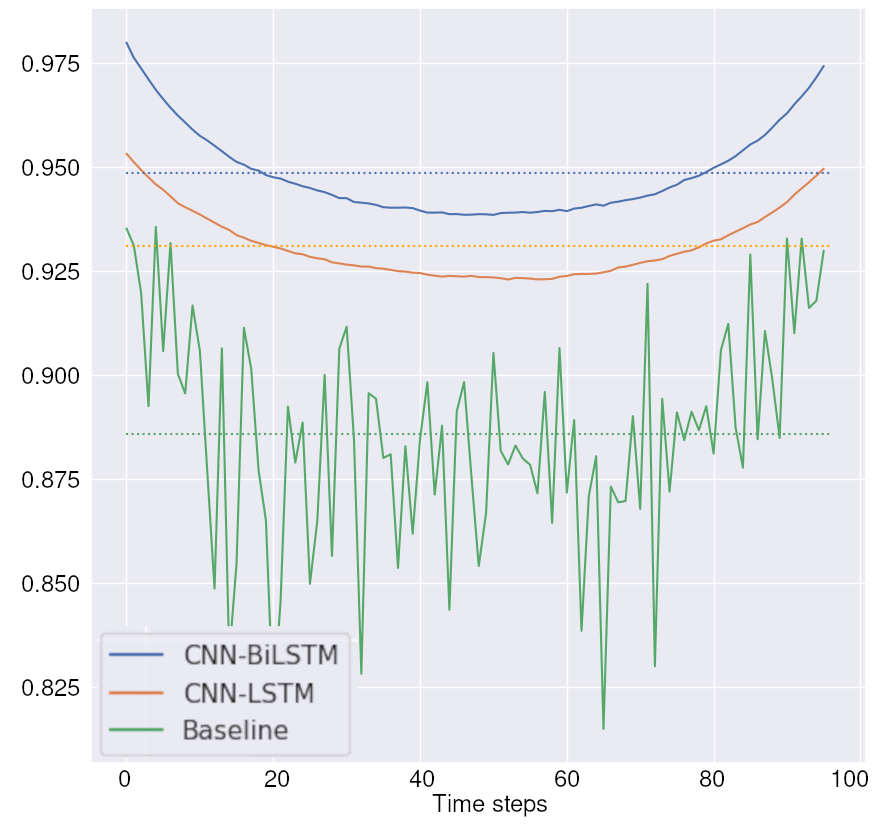}
    \caption{$R^2$ and their means ($\overline{R^2}$, dotted lines) for all models on BRE-CAG apartments. 
    }
    \label{fig:r2s_distrib_unseen}
\end{figure}

As done above for the tests on the SIM data, we inspect the $R^2$ of the three models on the BRE-CAG apartments for all the 96 time steps composing the 1-day predictions.
This is shown in Fig~\ref{fig:r2s_distrib_unseen}, from which we make the following observations:
\begin{itemize}
    \item Also in the case of unseen apartments, the CNN-BiLSTM achieves the best performances, both overall and in every single time step.
    \item Both CNN-BiLSTM and CNN-LSTM models achieve good results, showing a promising ability of generalizing to unseen data. In addition, both models show a better accuracy in the first and last part of the window (as expected), with results that are symmetric in time, thus suggesting a good balance in exploiting past and future data.
    \item In contrast to the case of SIM apartments (see Fig.~\ref{fig:r2s_distrib}), the Baseline model's values show very high variation in the whole sequence of 96 time steps, besides a lower average ($< 90\%$).
    This is the first evidence of a model working much worse than the other two.
\end{itemize}

%
%
To summarize, the results of the tests over both SIM and BRE-CAG apartments point out the following:

\begin{enumerate} \label{summary}
    \item All three models achieve low errors and relatively high $\overline{R^2}$, meaning that they all are effective enough for being used in a data-filling task. 
    Also, the MAPE values, that can be used for comparison with completely different works, range in very low levels and show how well the models approximate the real values.
    \item The CNN-BiLSTM model performs significantly better than the other two, showing in some cases almost half of their errors.
    \item With respect to the Baseline model, the CNN-LSTM has slightly higher errors in the case of SIM apartments, but way lower errors on data from the BRE-CAG apartments.
    For both SIM and BRE-CAG apartments, the CNN-LSTM model approximates the true values more realistically than the Baseline, better reproducing the fluctuations in the original data.
\end{enumerate}
Generally, the preferable method of filling fluctuating time series is a model with low errors but enough variation in its predictions, avoiding flat predictions around the mean of the real values. 
Therefore, the CNN-BiLSTM is the best performing among the three models, and the CNN-LSTM, despite the higher errors on the SIM apartments, is preferable to the Baseline model for its ability to generalize to unseen data. \label{concl_4}

\smallskip
\smallskip

To conclude this section, we provide some visual examples of predictions done by the three models.
This collection of examples includes some of the different situations that we encountered while visually inspecting the results, but it is not meant to be exhaustive due to the huge number of samples in our dataset.
Furthermore, these examples are not relevant for a thorough comparison of the models, since they show specific situations whose frequency in the samples has not been computed.
For a comprehensive assessment of the three models and their performances, indeed, we refer to the previous part of this section.

Each of the following figures consists of three sub-figures that show the real values (in orange) and their predicted approximations (in blue) from the three models respectively. 
In all figures, the left plot refers to the Baseline model, the middle one to the CNN-LSTM, and the right one to the CNN-BiLSTM.

\begin{figure}[t!]
    \centering
    \begin{subfigure}{0.3\linewidth}
        \centering
        \includegraphics[height=1.24in]{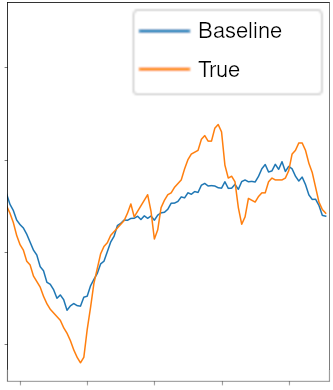}
    \end{subfigure}
    ~
    \begin{subfigure}{0.3\linewidth}
        \centering
        \includegraphics[height=1.25in]{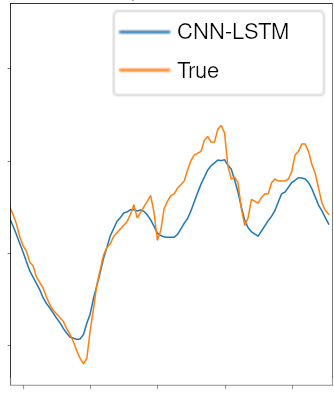}
    \end{subfigure}
    ~
    \begin{subfigure}{0.3\linewidth}
        \centering
        \includegraphics[height=1.25in]{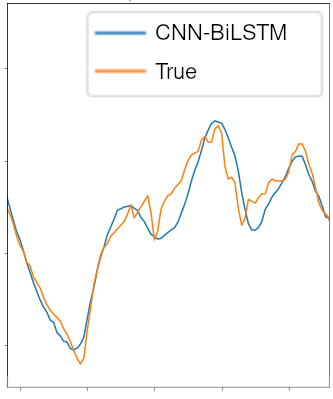}
    \end{subfigure}
    \caption{Predictions from the three models. Example~1, SIM apartments.}
    \label{fig:plots_EX1}
\end{figure}

\begin{figure}[t!]
    \centering
    \begin{subfigure}{0.3\linewidth}
        \centering
        \includegraphics[height=1.45in]{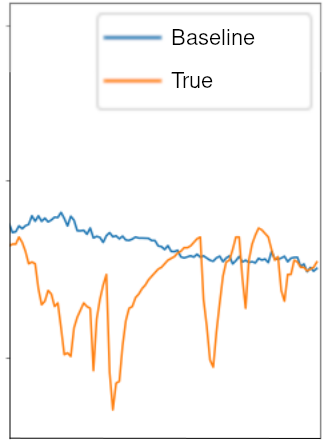}
    \end{subfigure}
    ~
    \begin{subfigure}{0.3\linewidth}
        \centering
        \includegraphics[height=1.45in]{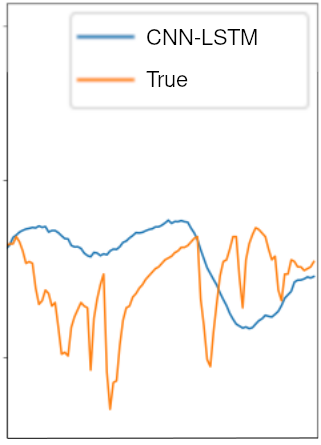}
    \end{subfigure}
    ~
    \begin{subfigure}{0.3\linewidth}
        \centering
        \includegraphics[height=1.45in]{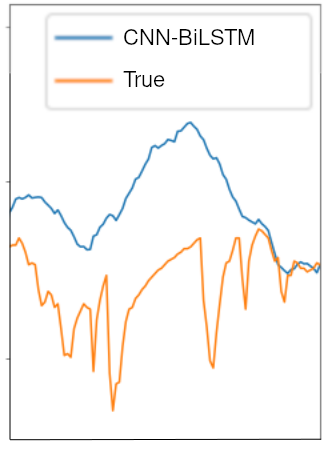}
    \end{subfigure}
    \caption{Predictions from the three models. Example~2, SIM apartments.}
    \label{fig:plots_EX2}
\end{figure}

\begin{figure}[t!]
    \centering
    \begin{subfigure}{0.3\linewidth}
        \centering
        \includegraphics[height=1.25in]{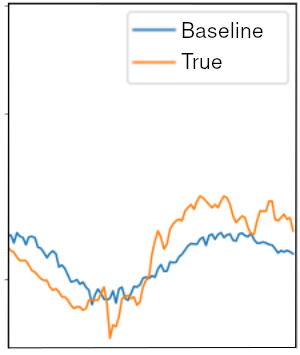}
    \end{subfigure}
    ~
    \begin{subfigure}{0.3\linewidth}
        \centering
        \includegraphics[height=1.25in]{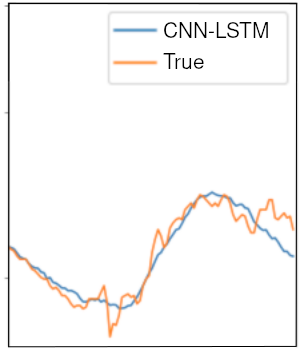}
    \end{subfigure}
    ~
    \begin{subfigure}{0.3\linewidth}
        \centering
        \includegraphics[height=1.25in]{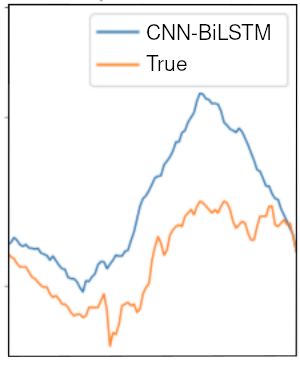}
    \end{subfigure}
    \caption{Predictions from the three models. Example~3, SIM apartments.}
    \label{fig:plots_EX5}
\end{figure}

Figures~\ref{fig:plots_EX1}, \ref{fig:plots_EX2}, \ref{fig:plots_EX5} refer to samples from the test set of the SIM apartments, which are the apartments used for the training of the models.
In the example shown in Fig.~\ref{fig:plots_EX1}, all three models achieve a relatively good approximation of the true values. While the Baseline's prediction are more flattened, the other two accurately estimate the real values, with the CNN-BiLSTM having the best fit.
Fig.~\ref{fig:plots_EX2} instead shows a case in which all three models fail to accurately predict a sample with frequent oscillations. 
Despite the poor results of this sample, we must point out how the Baseline provides the most flattened prediction, while the CNN-BiLSTM captures the general oscillation but it is affected by a positive bias in the first two thirds of the window.
Finally, Fig.~\ref{fig:plots_EX5} shows an example where the CNN-LSTM achieves the best approximation. While the Baseline seems close to the right trend but flattened, the CNN-BiLSTM also follows a relevant trend but is again positively biased.

\begin{figure}[t!]
    \centering
    \begin{subfigure}{0.3\linewidth}
        \centering
        \includegraphics[height=1.45in]{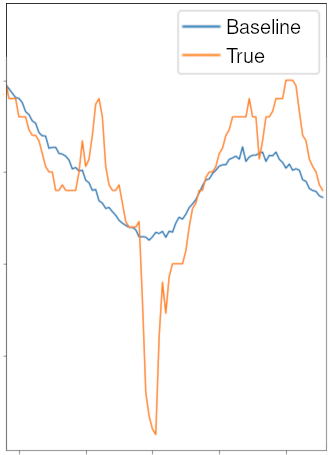}
    \end{subfigure}
    ~
    \begin{subfigure}{0.3\linewidth}
        \centering
        \includegraphics[height=1.45in]{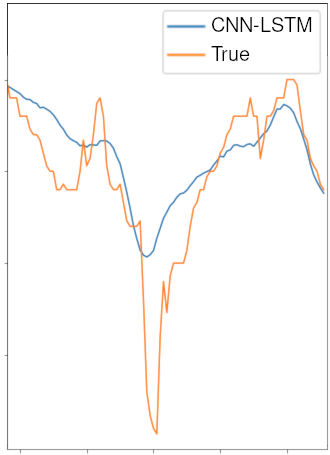}
    \end{subfigure}
    ~
    \begin{subfigure}{0.3\linewidth}
        \centering
        \includegraphics[height=1.45in]{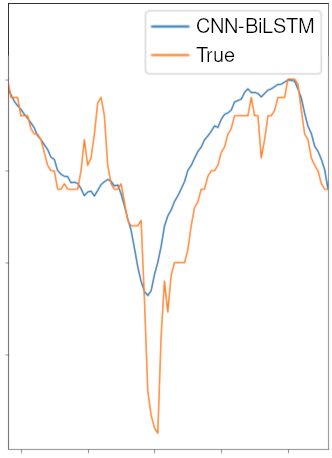}
    \end{subfigure}
    \caption{Predictions from the three models. Example~1, BRE-CAG apartments.}
    \label{fig:plots_EX1_U}
\end{figure}
\begin{figure}[t!]
    \centering
    \begin{subfigure}{0.3\linewidth}
        \centering
        \includegraphics[height=1.65in]{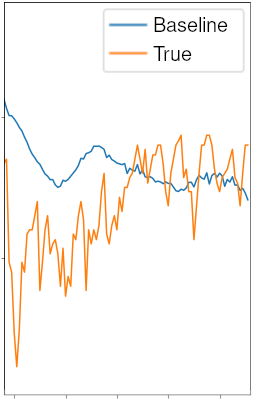}
    \end{subfigure}
    ~
    \begin{subfigure}{0.3\linewidth}
        \centering
        \includegraphics[height=1.65in]{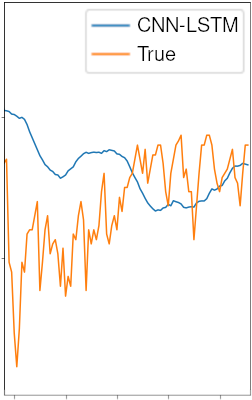}
    \end{subfigure}
    ~
    \begin{subfigure}{0.3\linewidth}
        \centering
        \includegraphics[height=1.65in]{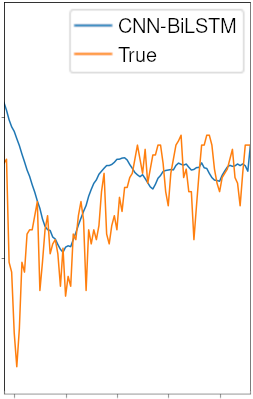}
    \end{subfigure}
    \caption{Predictions from the three models. Example~2, BRE-CAG apartments.}
    \label{fig:plots_EX3_U}
\end{figure}
\begin{figure}[t!]
    \centering
    \begin{subfigure}{0.3\linewidth}
        \centering
        \includegraphics[height=2.1in]{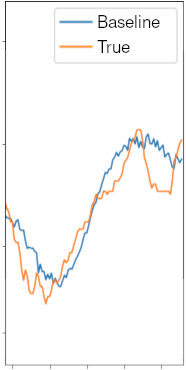}
    \end{subfigure}
    ~
    \begin{subfigure}{0.3\linewidth}
        \centering
        \includegraphics[height=2.1in]{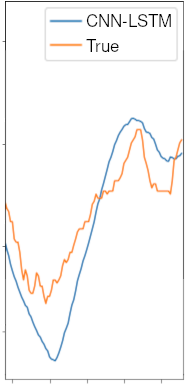}
    \end{subfigure}
    ~
    \begin{subfigure}{0.3\linewidth}
        \centering
        \includegraphics[height=2.1in]{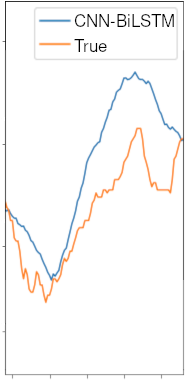}
    \end{subfigure}
    \caption{Predictions from the three models. Example~3, BRE-CAG apartments.}
    \label{fig:plots_EX4_U}
\end{figure}

Figures~\ref{fig:plots_EX1_U}, \ref{fig:plots_EX3_U}, \ref{fig:plots_EX4_U} are obtained instead from samples of the BRE-CAG apartments, which are unseen from the models during training.
Generally, the predictions in this case were significantly worse than those of the SIM apartments, as expected.
Fig.~\ref{fig:plots_EX1_U} shows an example where all three models are sufficiently good in capturing the overall trend of the real values. 
More in depth, while the Baseline does not capture the sudden drop around the mid of the window, the CNN-LSTM and CNN-BiLSTM seem to correctly grasp it, even if underestimating the drop.
Fig.~\ref{fig:plots_EX3_U} is instead an example in which all three models fail to estimate the real values and their trends. 
Nevertheless, only CNN-BiLSTM seems to be slightly outperforming the other two by creating some fluctuations that approach the curve of real values.
Lastly, Fig.~\ref{fig:plots_EX4_U} shows an example where the Baseline captures better the variation and the values of the true time series, while the other two models show a slightly worse behavior, but still capture the main trend.

%
%

\section{Conclusions} \label{Conclusion and future work}
%
In general, the results outlined in the previous section demonstrate the effectiveness of all three models --- the two brand new ones presented in this paper plus the one from~\cite{song2020time} taken as a baseline --- in achieving good data filling results.
In detail, both models we developed work substantially better than the Baseline, especially when generalizing to totally unseen data. In particular, the high performances obtained by the CNN-BiLSTM model can be seen as the main outcome of this work.
Moreover, the fact that the Baseline achieves sufficient results on our dataset demonstrates the robustness of the LSTM approach and of our choice to use this model as a comparison for our work.

%
An important difference between our two models and the Baseline is that the CNN-LSTM and the CNN-BiLSTM make use of external data (the TEXT sensor) in order to predict the target time series (the TEM sensor). This is not the case of the Baseline, that uses only data from the TEM time series in order to forecast the missing values. This disparity contributed probably in an essential way to the higher performances of our models, especially when testing on data coming from unseen apartments (the ones of BRE-CAG, see Table~\ref{table:errors_unseen}). The other major difference between our work and the Baseline is the presence of CNN layers in our models. The fact that combinations of CNN and LSTM layers show more promising results than purely LSTM networks was suggested already by~\cite{lara2021experimental} and finds confirmation in our work.
Nevertheless, we must point out that a precise quantification of the added value given by CNN layers and the use of external data would require further investigation and it is outside the scope of this work.

%
Restricting instead the focus on our two models, they both make use of the same input data and they differ only in the way they handle and combine the information preceding and following the data gap to be filled. Indeed, the CNN-LSTM model is made of two networks containing CNN and LSTM layers that estimate the data gap using, respectively, the previous and the following 6 days. Both networks are trained and, after training, their forecasts are combined using the sigmoid function --- see~\eqref{sigmoid_pred} --- obtaining the final output of the CNN-LSTM model. On the contrary, the CNN-BiLSTM is made of one single network that exploits the power of BiLSTM layers to process the data from the previous and the following 6 days in both directions. The main difference is thus in the fact that the latter model learns automatically, during the training phase, how to combine past and future data in a single forecast for the missing values. From our work, we can conclude that this "black box" strategy learnt by the BiLSTM layer is more effective than the explicit combination of forward and backwards previsions done with formula~\eqref{sigmoid_pred}. In addition, considering the performances of the separate CNN-LSTM-Onwards and CNN-LSTM-Backwards networks, we tend to exclude that a combination different than~\eqref{sigmoid_pred} can lead to overcome the results achieved by the CNN-BiLSTM model.

%
Concerning instead the size of the data gaps, we decided to keep the same one for the whole work in order to focus on the differences among the models. Besides, the size of the gap we chose (1 day) is large enough to require a more sophisticated approach than simple interpolation, but it is short enough to maintain some predictability.
Nevertheless, it would be interesting to assess the proposed models on gaps of different lengths, in particular for what concerns the most promising CNN-BiLSTM model. 
For the case of shorter time scales, such as few hours, it would be worth comparing this kind of LSTM-based models with classical statistical methods. Indeed, when the oscillations in the data are small, the advantage of using DL techniques might be reduced.
Regarding longer time scales instead, we point out that a combination of the models we developed could be used to fill gaps longer than one day, even if this was not the scope of the present work. 
Indeed, the CNN-LSTM-Onwards and the CNN-LSTM-Backwards (which are the two networks composing the CNN-LSTM model) can be used to predict missing data using, respectively, only the past and future samples. In this way, a larger gap can be progressively reduced until it reaches the size of 1 day and then the CNN-BiLSTM model can be applied. Naturally, as pointed out also in~\cite{contractor2021efficacy}, this approach is highly exposed to accumulate the error.

%
Finally, despite of the good results achieved in general by the models we propose, we must point out that the visual inspection of some particular cases at the end of Section~\ref{Errors comparison} shows large margins for improvement of the forecasts.
Being aware that the randomness in the data prevents the models from achieving extremely high accuracy, using other inputs in addition to TEM and TEXT could lead to substantial advancements. 
Some possibilities are heating, cooling, or ventilation time series, which have a direct effect on the temperature.
Using as additional input the CO$_2$ levels, which are correlated with the number of people present inside the apartment, or the relative humidity, is also an interesting option. In our case, this was prevented by the fact that one sensor unit captures, at the same time, the internal temperature, the relative humidity, and the CO$_2$ concentration, which means that all three values are lost in case of failure of the sensor.
The best way to exploit additional features would probably be to do feature engineering to obtain the input for the DL model. This would open up several possibilities on how to combine features and could potentially lead to substantial improvement in the results.
Moreover, the hyper-parameters of the models were decided manually after several experiments while the training and validation metrics were monitored. The model selection phase could be strengthened by following automated procedures, such as Grid-Search or Hyper-parameters Optimization Algorithms, which allow assessing more options for the hyper-parameters.
Finally, embedding physical principles in the model would be another promising approach. Specifically, two possibilities would be inserting equations into the neural network or creating an interface between the DL model and a physical model of the apartments.

%
%

\section*{Acknowledgments}

The authors would like to thank Mouna Kacimi and Matteo Ceccarello for fruitful discussions on the topic of this paper and for providing valuable input to this work. 

The research leading to these results has received funding from the European Union’s Seventh Program for research, technological development and demonstration under grant agreement No. 609019 within SINFONIA project. The European Union is not liable for any use that may be made of the information contained in this document which is merely representing the authors view.

%
%
\printbibliography

\end{document}